\documentclass{article} %
\usepackage{iclr2027_conference,times}

\definecolor{cvprblue}{rgb}{0.37,0.59,0.79}
\definecolor{cornellred}{rgb}{0.7, 0.11, 0.11}
\usepackage[pagebackref,breaklinks,colorlinks,citecolor=cvprblue,linkcolor=cornellred]{hyperref}
\usepackage{url}
\usepackage{graphicx}
\usepackage{caption}
\usepackage{booktabs}
\usepackage{amsmath}
\usepackage{amssymb}
\usepackage[capitalize]{cleveref}
\usepackage{array}
\usepackage{tabularx}
\usepackage{adjustbox}
\usepackage{booktabs}
\usepackage{tikz}

\crefname{figure}{Figure}{Figures}
\crefname{table}{Table}{Tables}
\crefname{section}{Section}{Sections}
\crefname{subsection}{Section}{Sections}
\crefname{equation}{Equation}{Equations}
\crefname{appendix}{Appendix}{Appendices}

\newcommand{\red}[1]{{\color{red}#1}}
\definecolor{kellygreen}{rgb}{0.3, 0.73, 0.09}
\newcommand{\greencap}[1]{\textcolor{kellygreen}{#1}}

\usepackage{multirow}

\newcommand{\videoInput}{\mathbf{I}}
\newcommand{\matmap}{\mathbf{mat}}
\newcommand{\diffusionModel}{\mathbf{f}}
\newcommand{\diffusionModelParams}{\theta}
\newcommand{\diffusionModelFn}{\diffusionModel_{\diffusionModelParams}}
\newcommand{\vaeEncoder}{\mathcal{E}}
\newcommand{\vaeDecoder}{\mathcal{D}}
\newcommand{\typeEmb}{\mathbf{c}_{\text{prompt}}}
\newcommand{\dataDistribution}{p_{\text{data}}}
\newcommand{\diffusionNoise}{\mathbb{\epsilon}}

\newcommand{\sota}[1]{%
  \textbf{#1}%
}

\newcommand{\subsota}[1]{%
    \underline{#1}%
}

\usepackage{pifont}

\newcommand{\cmark}{\ding{51}}
\newcommand{\xmark}{\ding{55}}

\newcommand{\redcross}{\red{\xmark}}
\newcommand{\greencheck}{\greencap{\cmark}}

\usepackage{listings}
\usepackage{xcolor}

\definecolor{codegreen}{rgb}{0,0.6,0}
\definecolor{codegray}{rgb}{0.5,0.5,0.5}
\definecolor{codepurple}{rgb}{0.58,0,0.82}
\definecolor{backcolour}{rgb}{0.95,0.95,0.92}

\lstdefinestyle{mystyle}{
    backgroundcolor=\color{backcolour},   
    commentstyle=\color{codegreen},
    keywordstyle=\color{magenta},
    numberstyle=\tiny\color{codegray},
    stringstyle=\color{codepurple},
    basicstyle=\ttfamily\footnotesize,
    breakatwhitespace=false,         
    breaklines=true,                 
    captionpos=b,                    
    keepspaces=true,                 
    numbers=left,                    
    numbersep=5pt,                  
    showspaces=false,                
    showstringspaces=false,
    showtabs=false,                  
    tabsize=2
}

\title{Texture Space Material Diffusion}

\author{Jacob Munkberg\\
NVIDIA
\And
Peter Kocsis \\
NVIDIA\\
\And
Jon Hasselgren \\
NVIDIA\\
}

\iclrfinalcopy %
\begin{document}

\newcommand{\figSystem}{
\begin{figure*}
    \centering
        \includegraphics[width=0.99\textwidth]{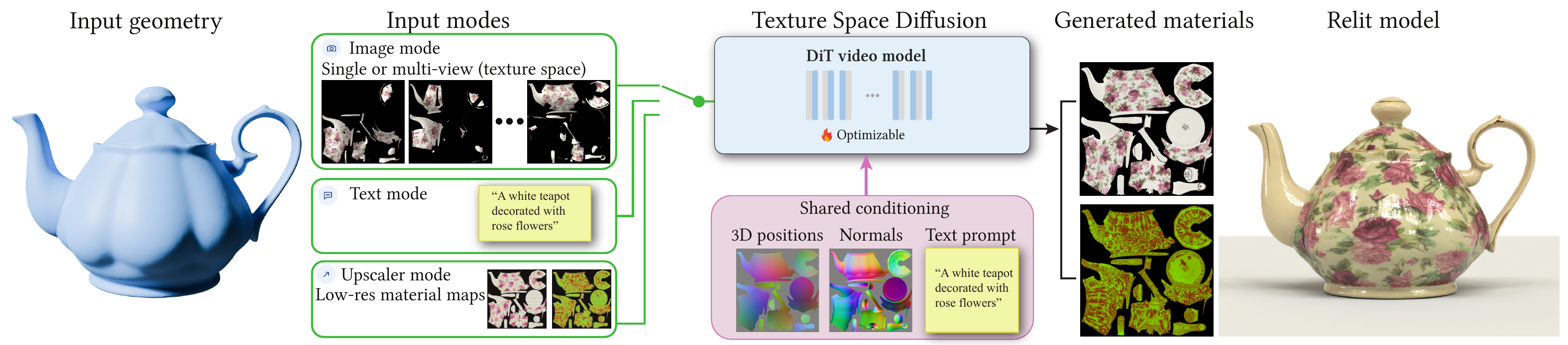}
    \vspace*{-2mm}
    \caption{\textbf{Method Overview.} Our pipeline leverages diffusion transformers in texture space to synthesize 
    high quality materials. We support multiple input modalities: 1) one or more views projected into texture space, which can be photographs, rendered images, or frames generated by diffusion models 2) a text prompt describing the material
    3) low-resolution material maps. In all cases, we condition the model on world space positions, normals and a text prompt.
    Our finetuned DiT successfully demodulates the (unknown) lighting, and creates clean albedo, roughness, and metallicity maps with high frequency details.
    The materials can be directly applied in standard content creation tools. %
    }
    \label{fig:system}
\end{figure*}
}

\newcommand{\figTexSpace}{
\begin{figure}
\centering
\small
    \includegraphics[width=0.99\columnwidth]{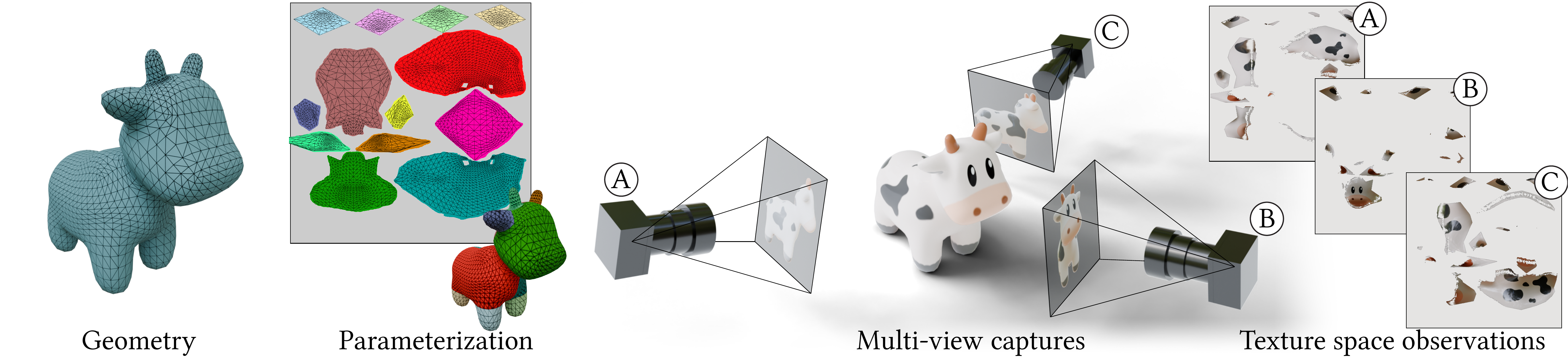}
\vspace*{-2mm}
\caption{\textbf{Texture Space Observations.}
    We leverage known geometry with a non-overlapping texture parameterization. 
    Given camera poses and geometry, we project each view back into texture space, which results in partially covered texture space observations. By formulating our diffusion process in the 2D texture space domain, we can leverage powerful image- and video diffusion model priors to synthesize high quality materials. 
}
\label{fig:texspace}
\end{figure}
}

\newcommand{\figCasual}{
\begin{figure}
    \centering
    \setlength{\tabcolsep}{1pt}
    \begin{tabular}{cccc}
       \includegraphics[width=0.24\columnwidth]{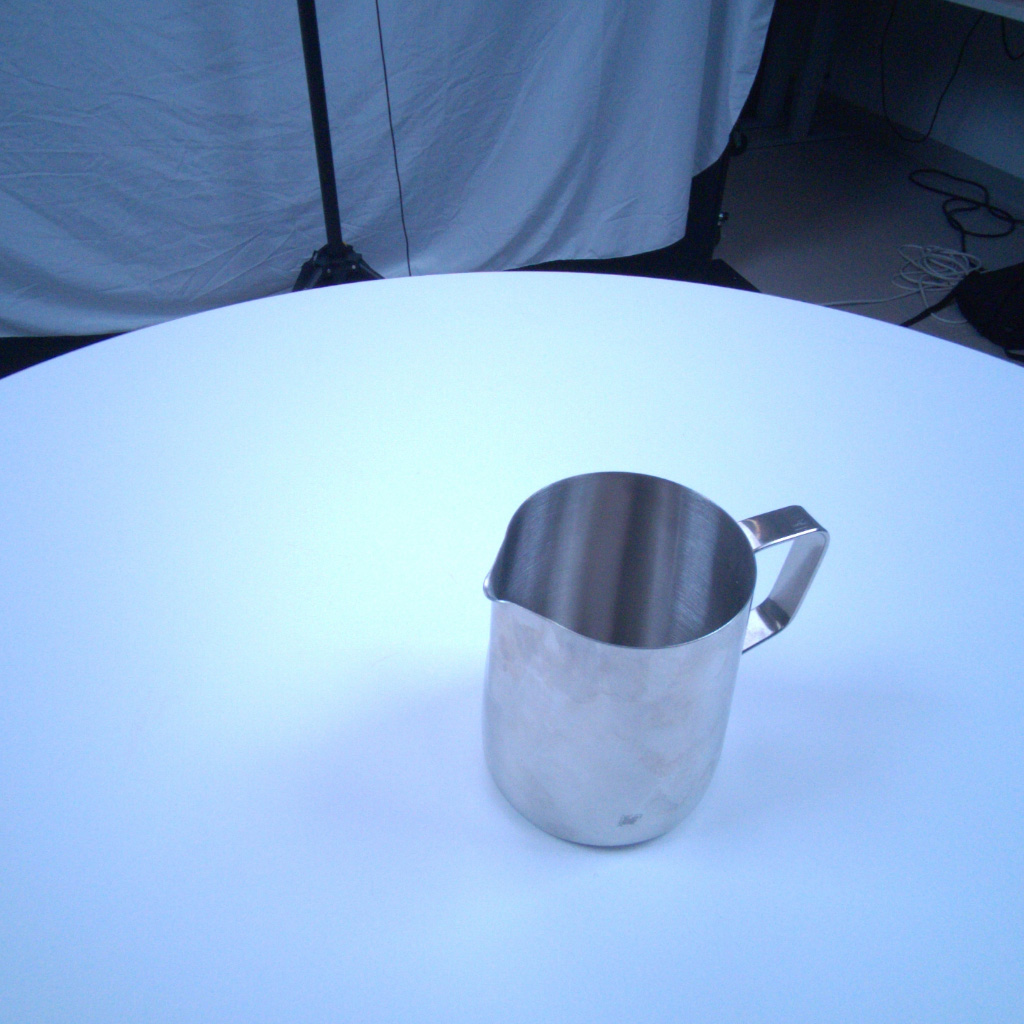} &
       \includegraphics[width=0.24\columnwidth]{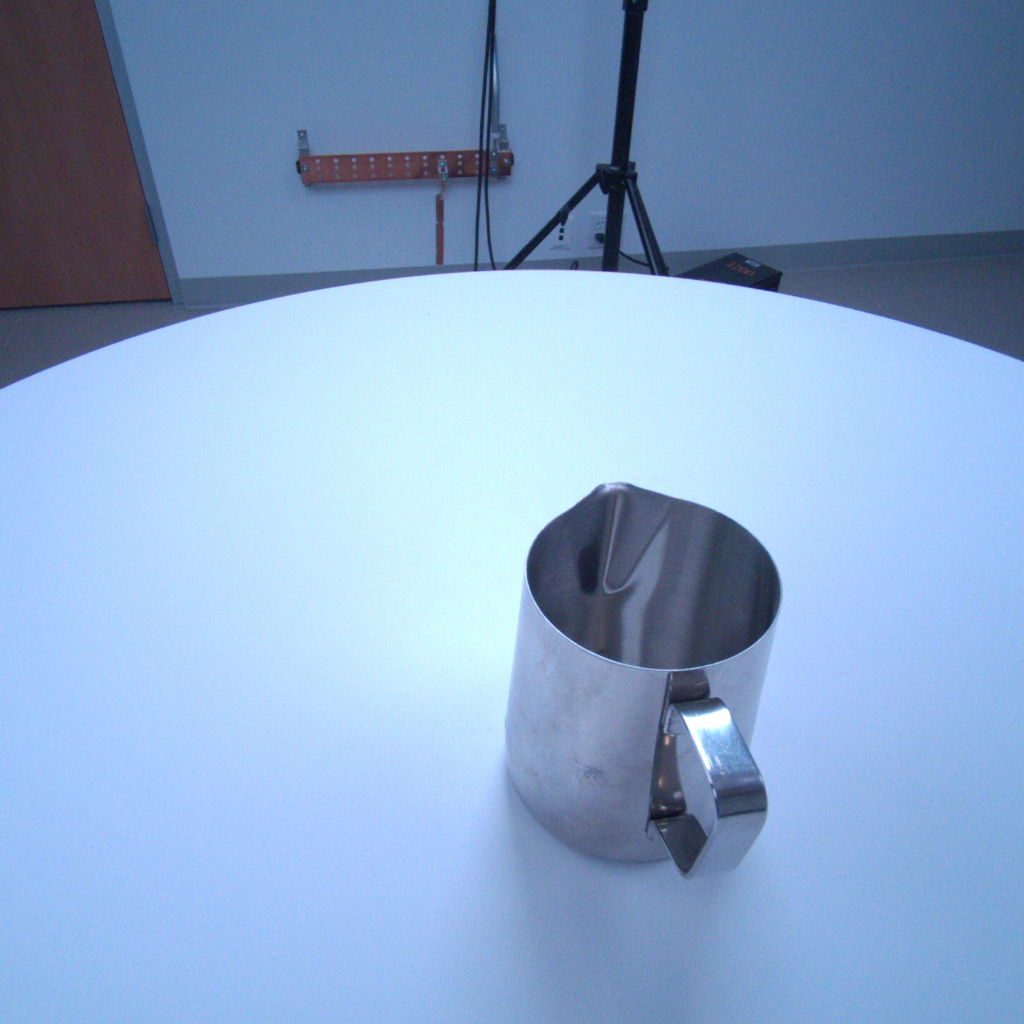} &
       \includegraphics[width=0.24\columnwidth]{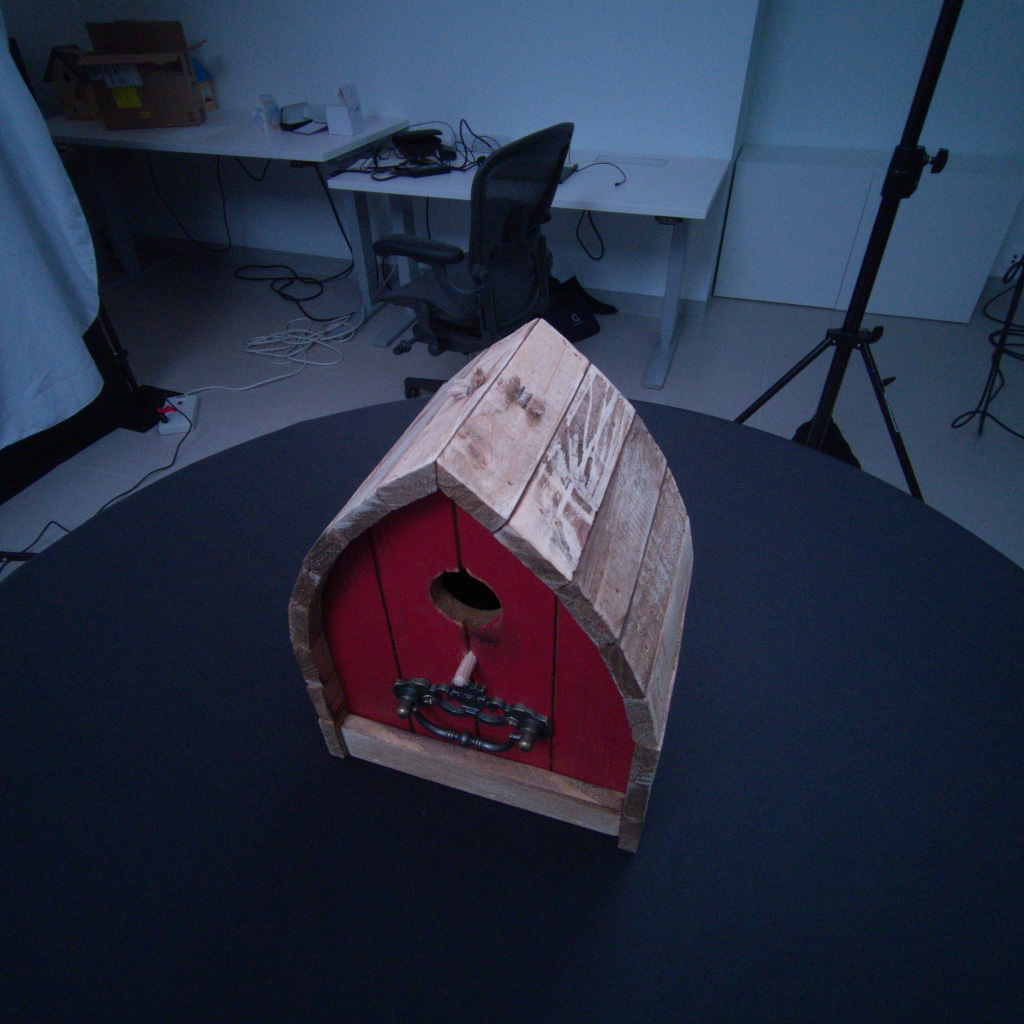} &
       \includegraphics[width=0.24\columnwidth]{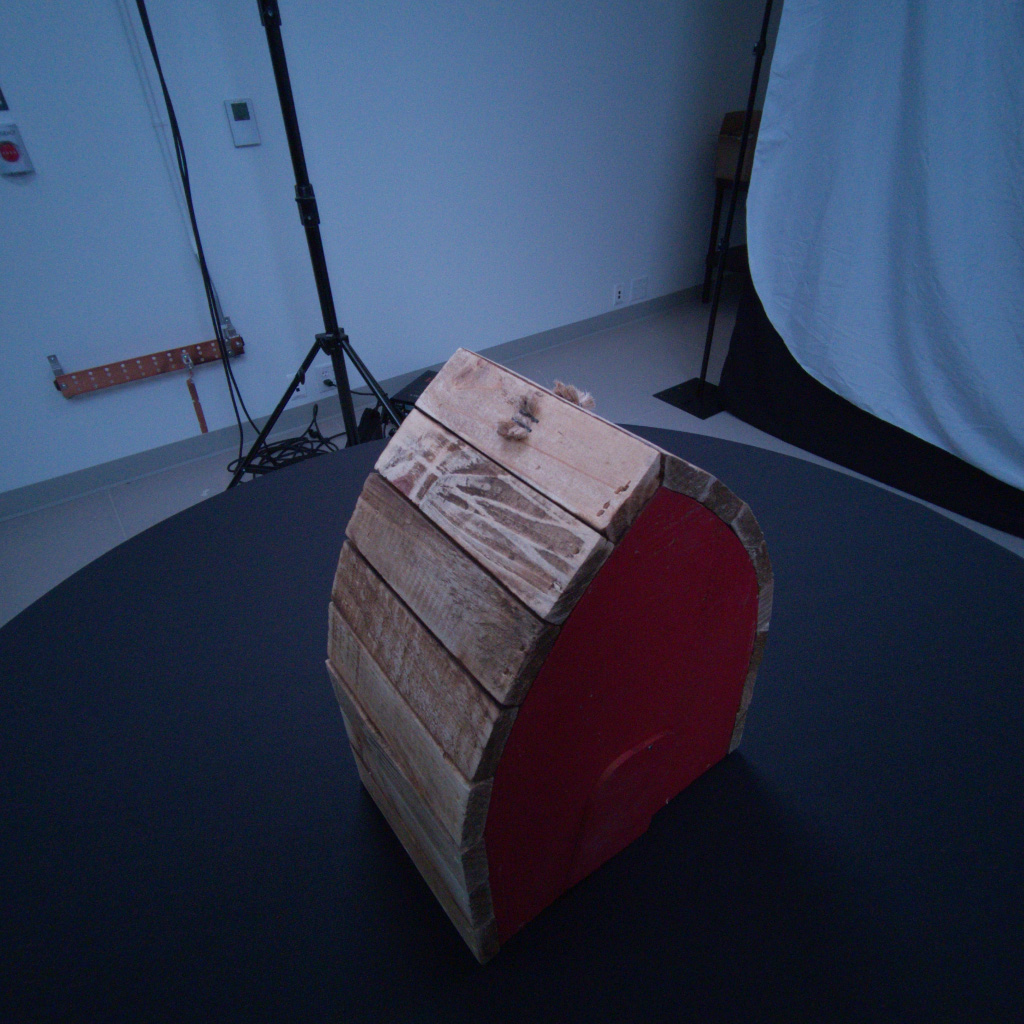} \\
       \includegraphics[width=0.24\columnwidth]{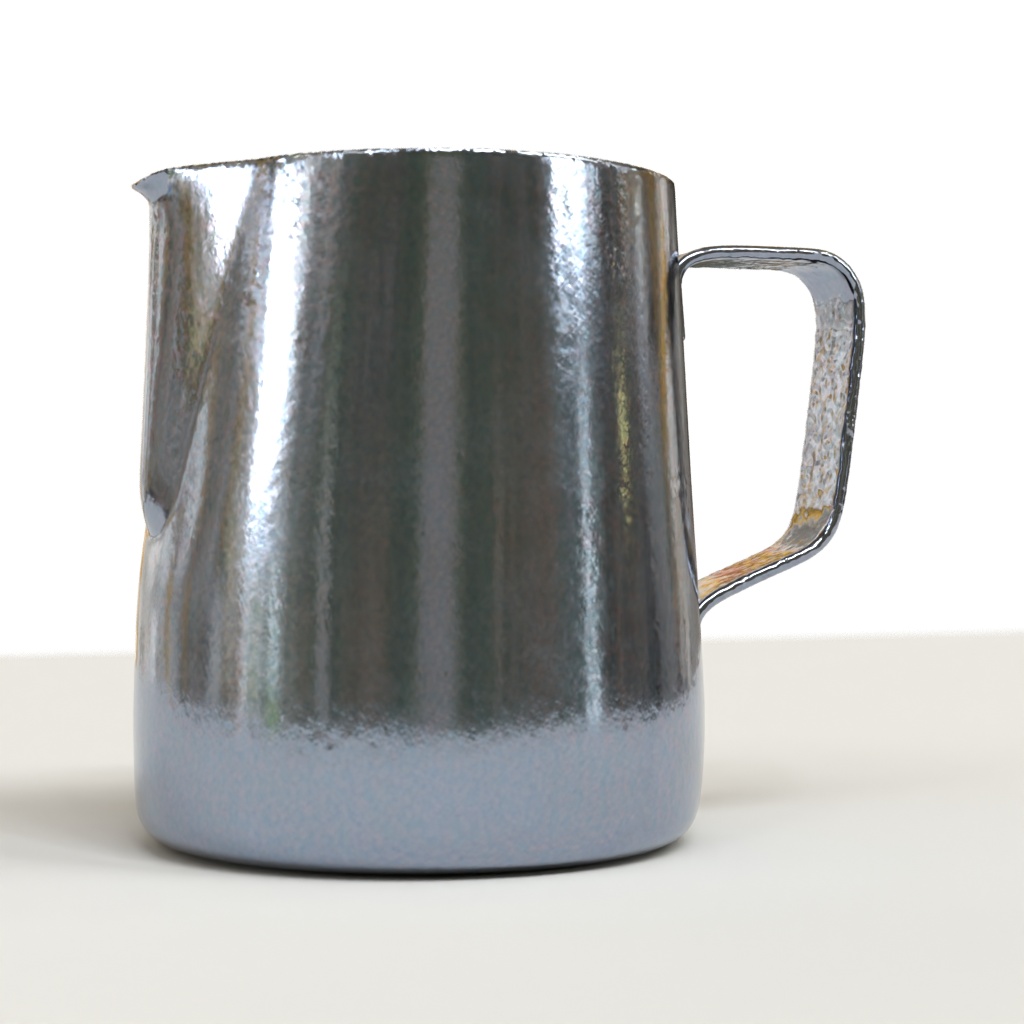} &
       \includegraphics[width=0.24\columnwidth]{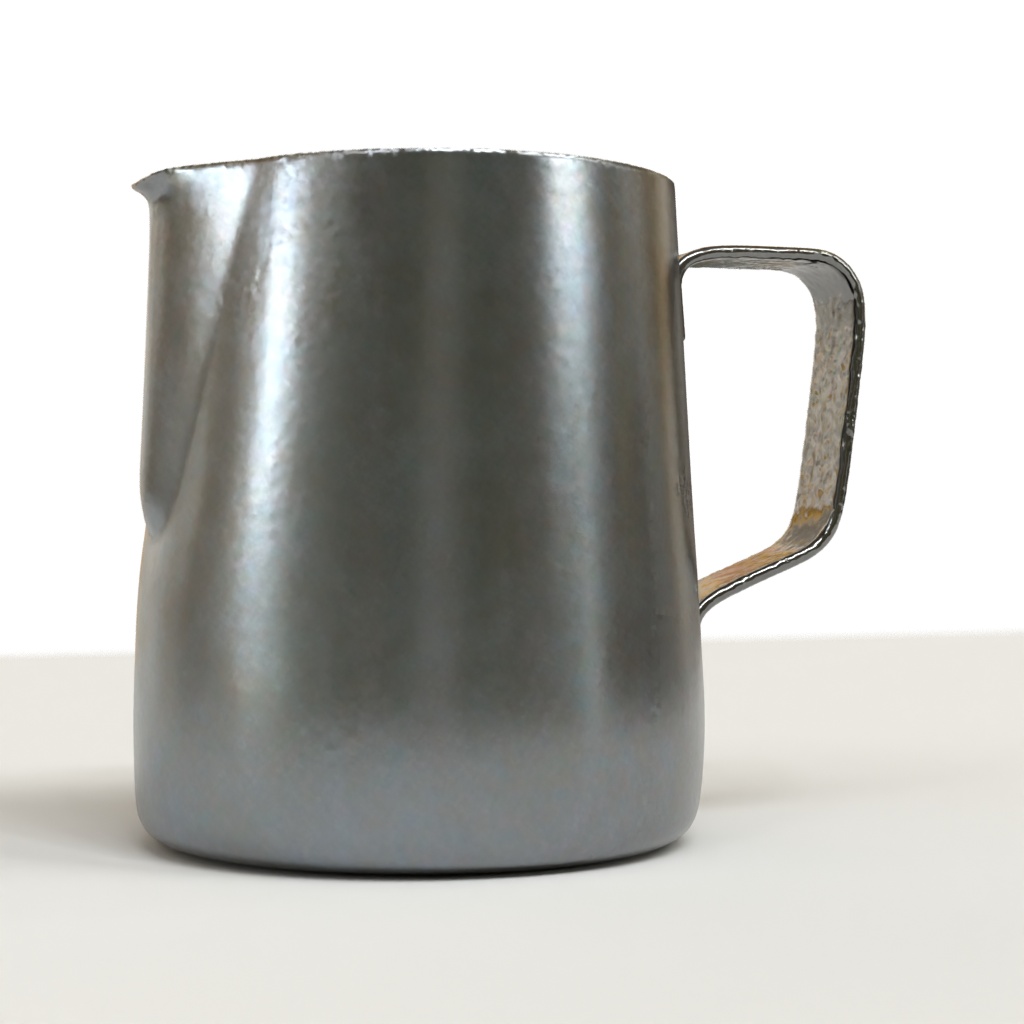} &
       \includegraphics[width=0.24\columnwidth]{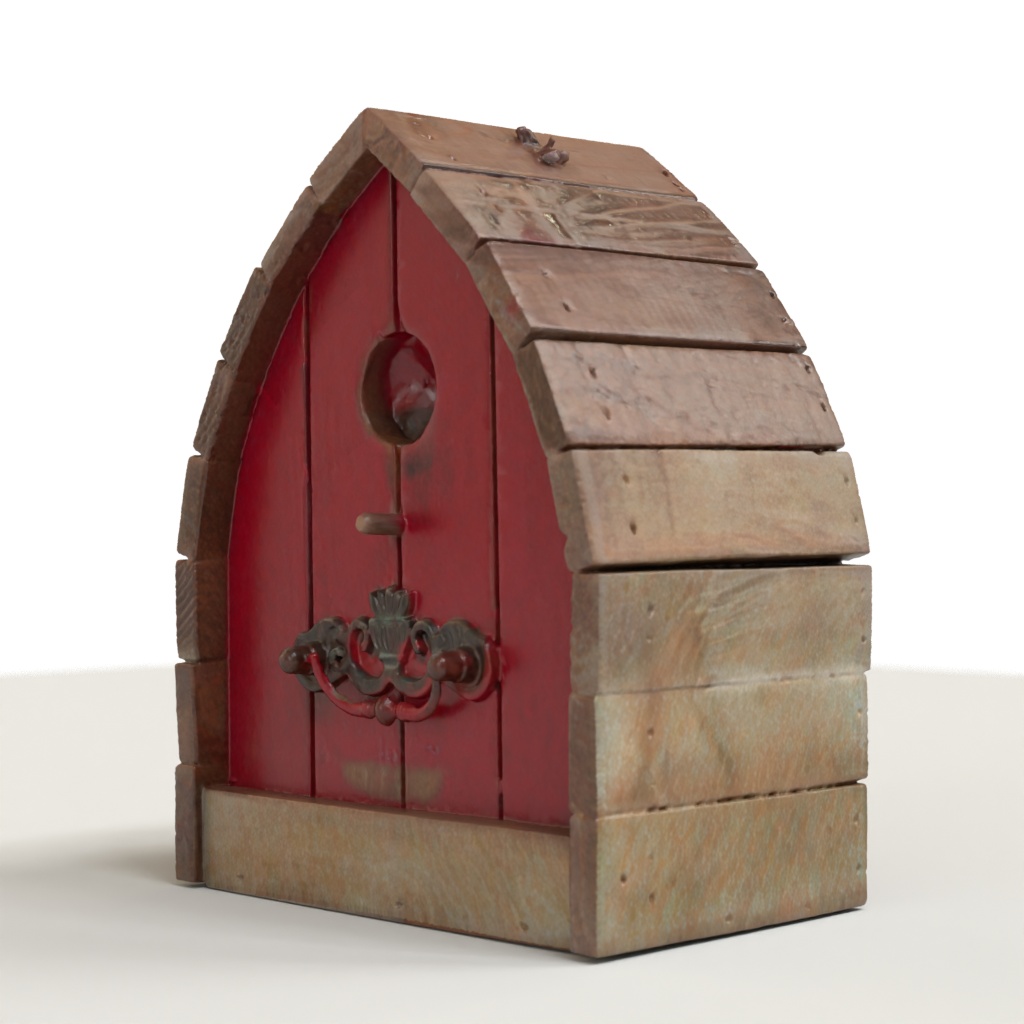} &
       \includegraphics[width=0.24\columnwidth]{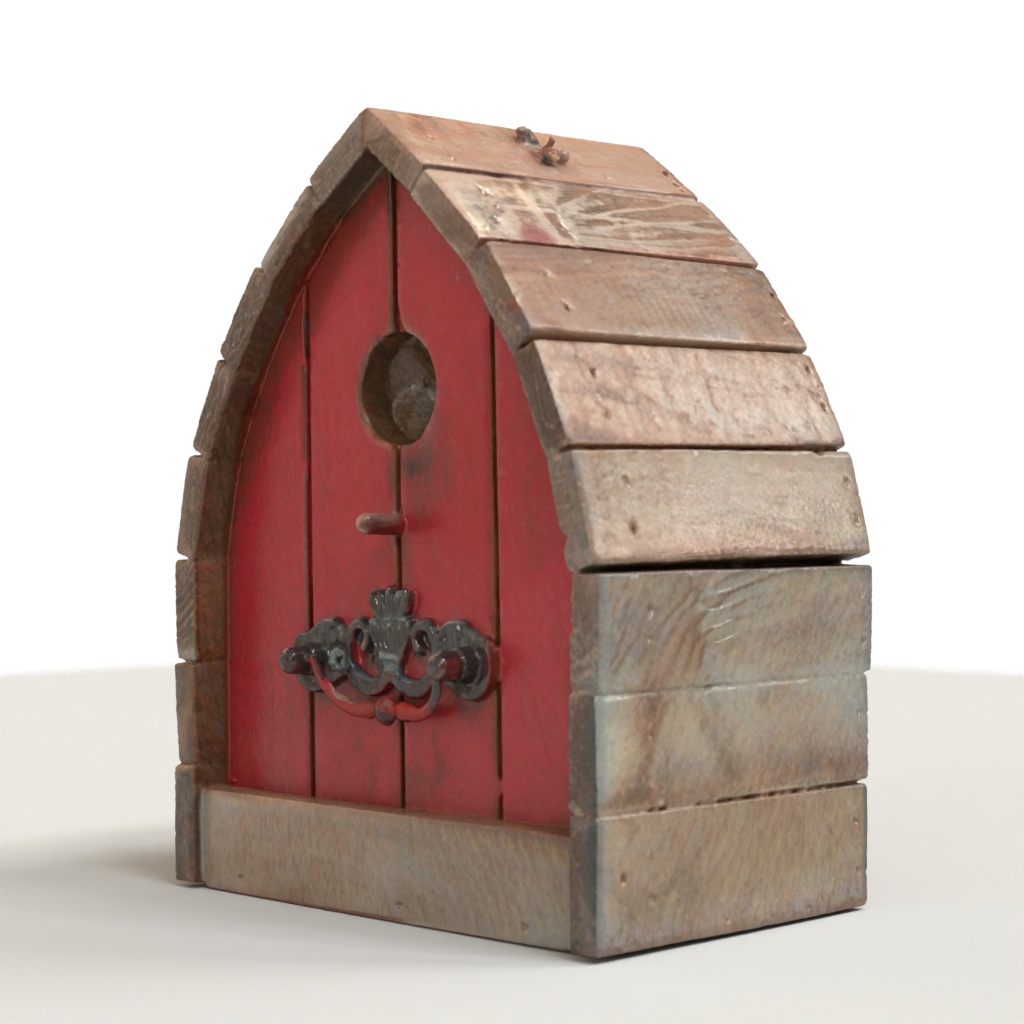} \\
    \end{tabular}
    \caption{Material generations from casually captured photographs, using 17 views from the DTC dataset~\cite{dong2025dtc}.
    We expect known camera poses and geometry in this test. 
    \textbf{Top}: Input views (two selected views per example).
    \textbf{Bottom}: Our extracted materials (two seeds per example) re-rendered in novel lighting in Blender.\vspace{1em}}
    \label{fig:casual}
 \end{figure}
}

\newcommand{\figRoPE}{
\begin{figure*}
    \centering
    \small
    \setlength{\tabcolsep}{3pt}
    \begin{tabular}{cc}
      Wan 2.1 RoPE & 3D-aware RoPE \\
      \includegraphics[width=0.47\textwidth]{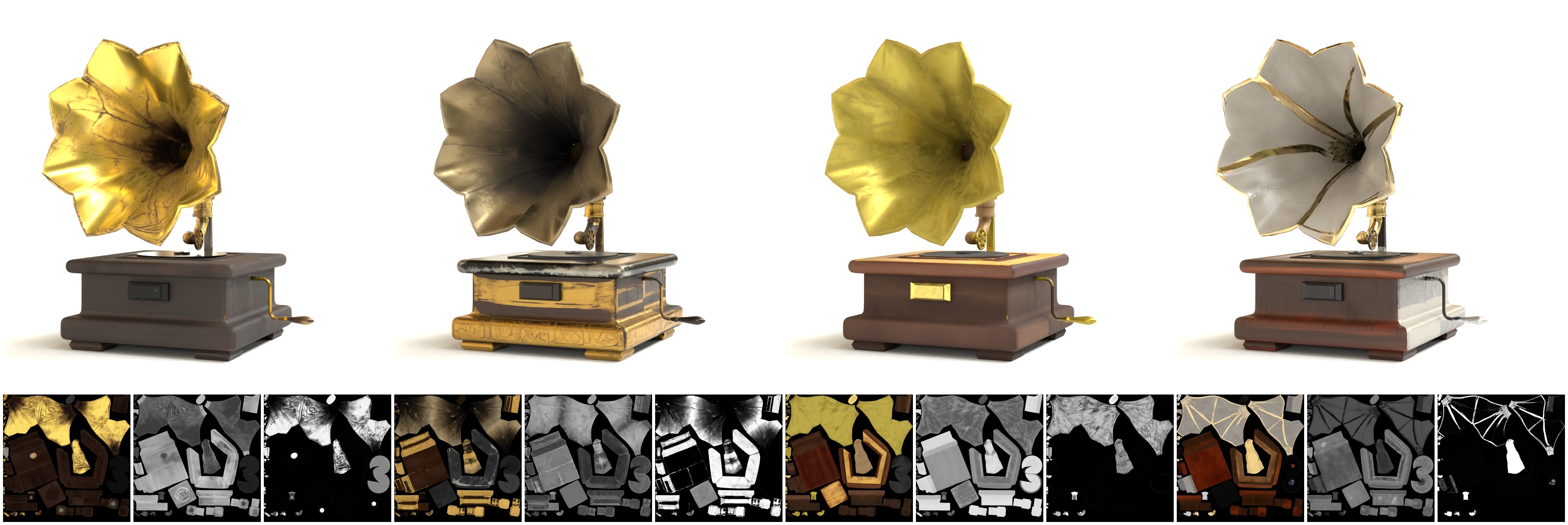} &
      \includegraphics[width=0.47\textwidth]{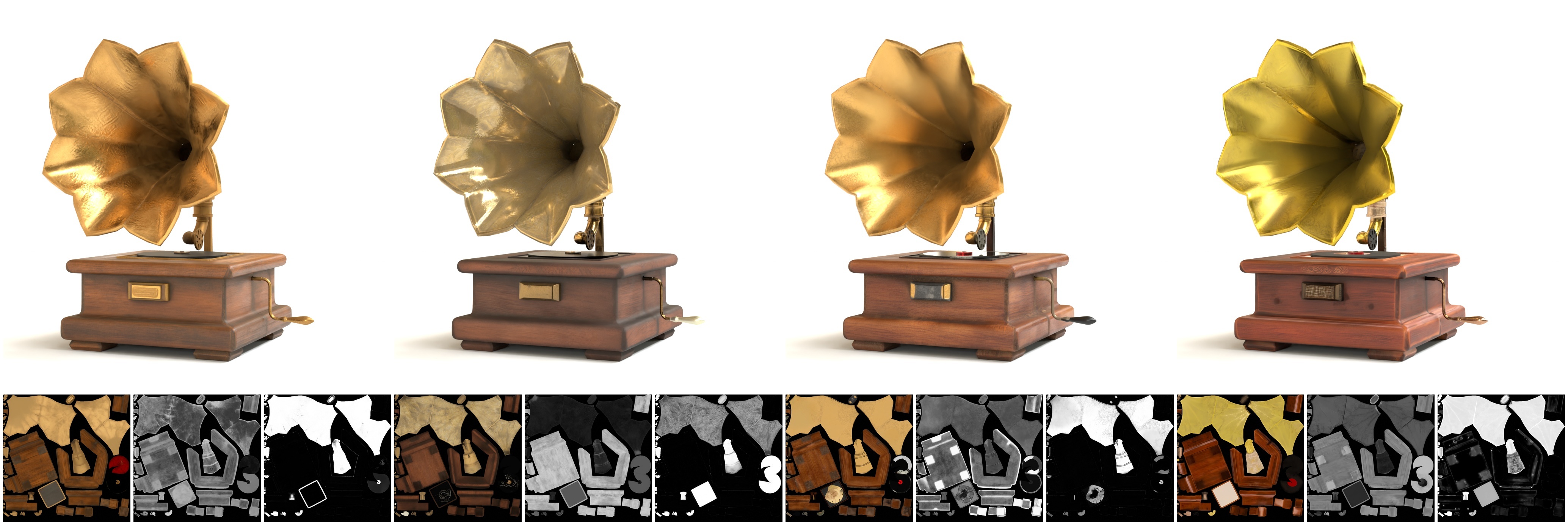}\\
    \end{tabular}
    \caption{\textbf{Stability.} Impact of RoPE embedding on a single text to 
    material example with varying random seed. As can be seen, the 3D-aware RoPE encoding 
    better captures the global structure of the object, and produces more consistent 
    materials. The small insets show the base color, roughness, and metallicity maps 
    for each generation.}
    \label{fig:figRoPE}

\end{figure*}
}

\newcommand{\figRoPEAttnSup}{
\begin{figure*}
    \centering
    \small
    \setlength{\tabcolsep}{1pt}
    \begin{tabular}{ccccc}
    \rotatebox[origin=c]{90}{Query A} &
    \raisebox{-0.5\height}{\includegraphics[height=0.226\textwidth]{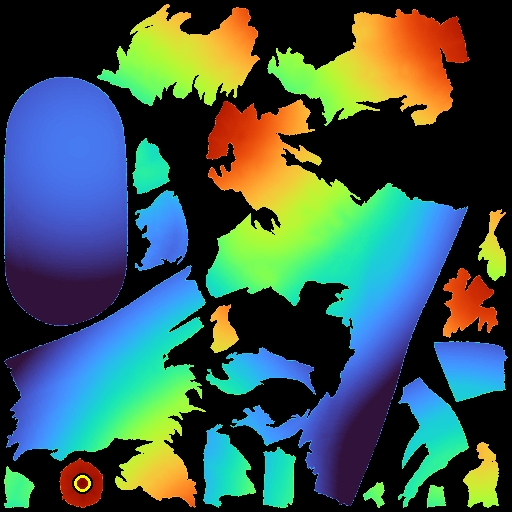}} &
    \raisebox{-0.5\height}{\includegraphics[height=0.226\textwidth]{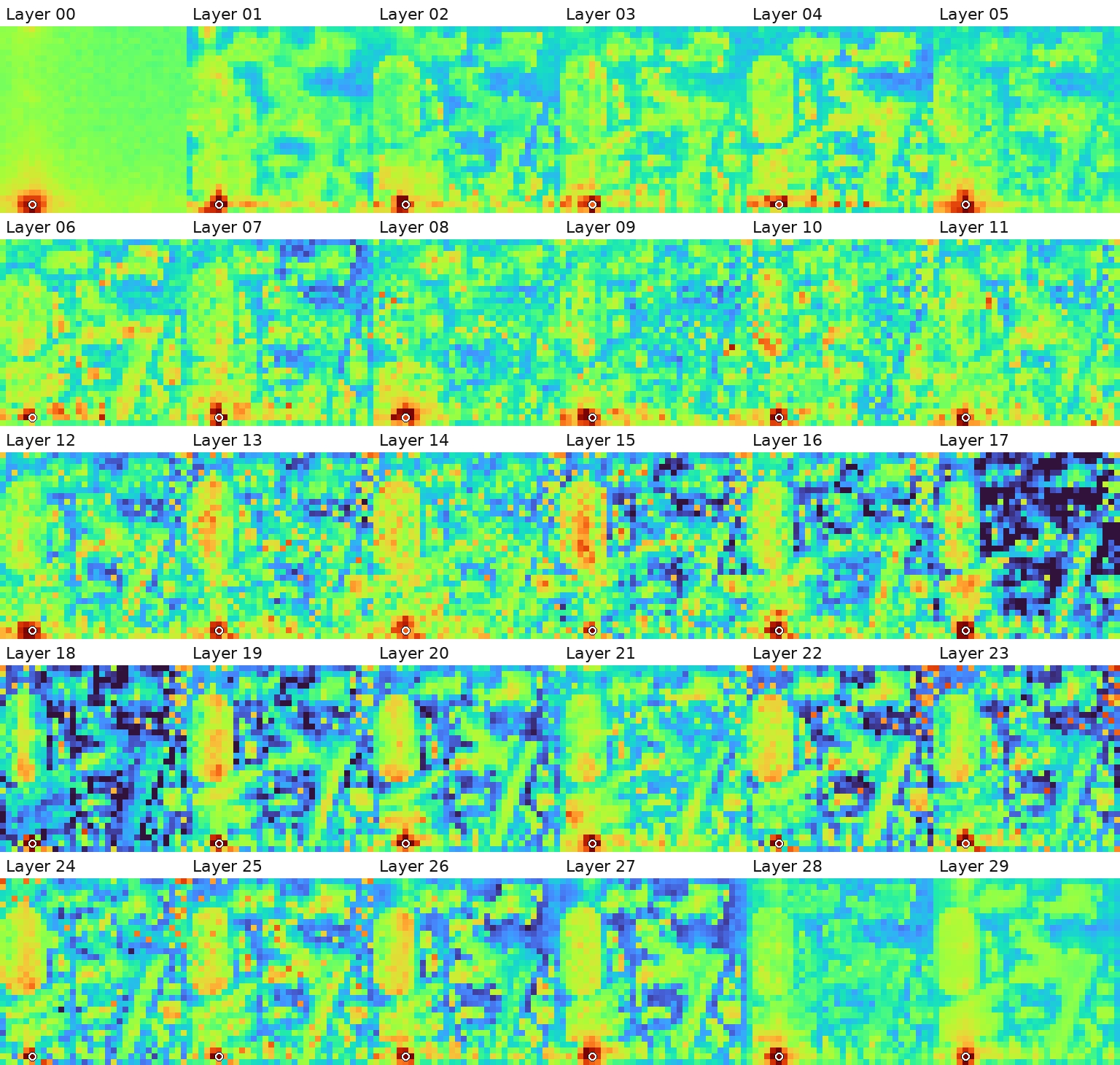}} &
    \raisebox{-0.5\height}{\includegraphics[height=0.226\textwidth]{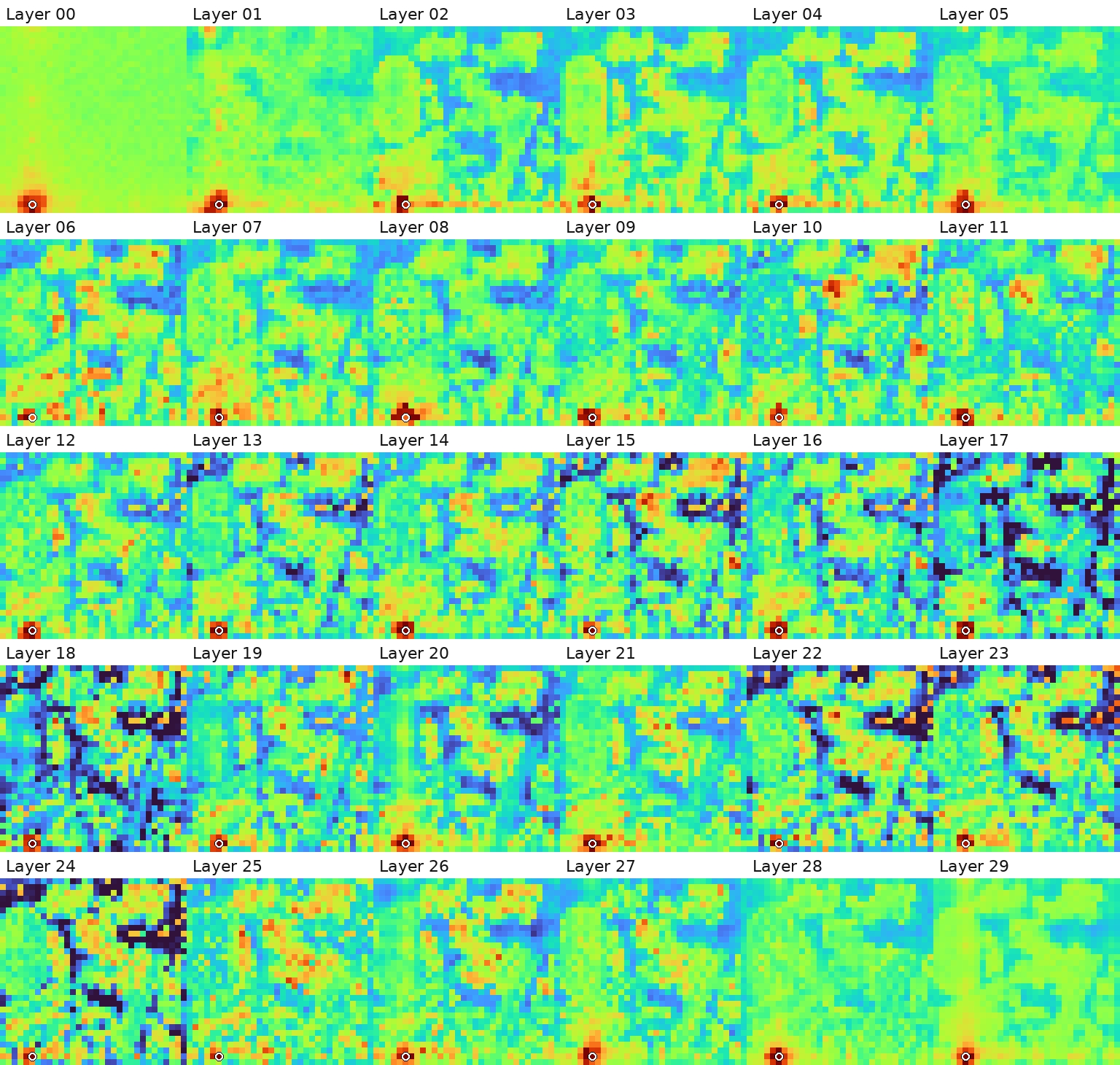}} &
    \raisebox{-0.5\height}{\includegraphics[height=0.226\textwidth]{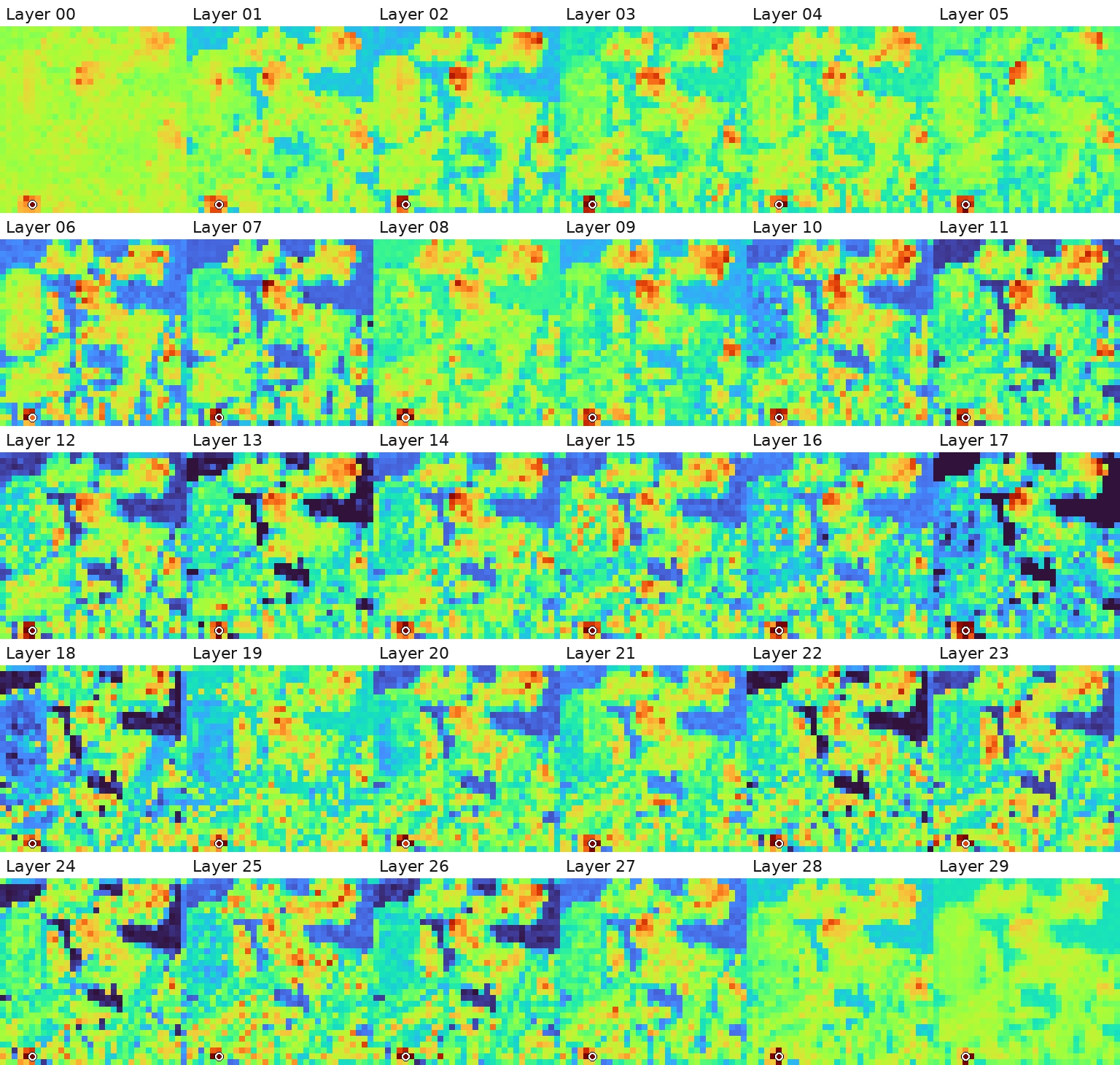}} 
    \\ \addlinespace[1pt]
    \rotatebox[origin=c]{90}{Query B} &
    \raisebox{-0.5\height}{\includegraphics[height=0.226\textwidth]{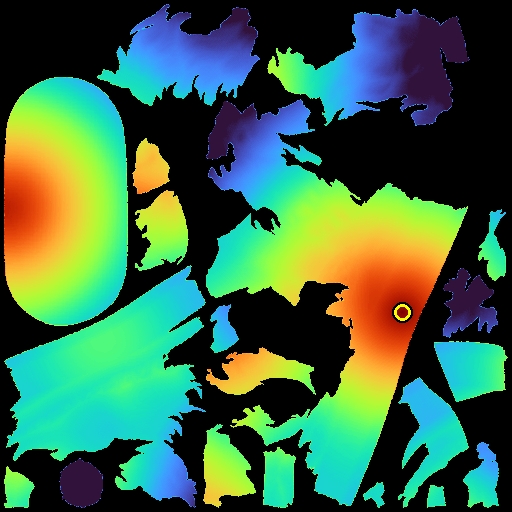}} &
    \raisebox{-0.5\height}{\includegraphics[height=0.226\textwidth]{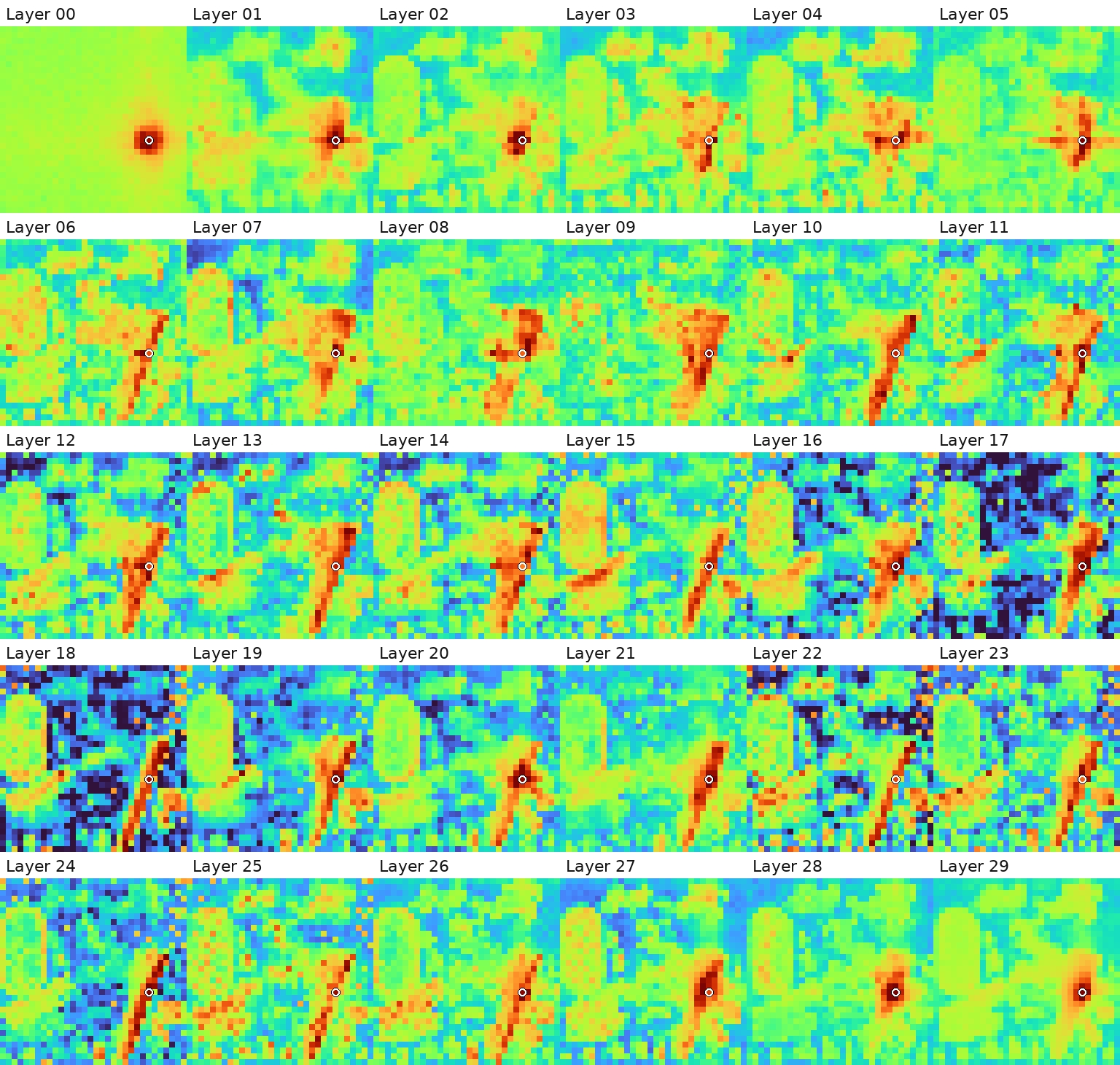}} &
    \raisebox{-0.5\height}{\includegraphics[height=0.226\textwidth]{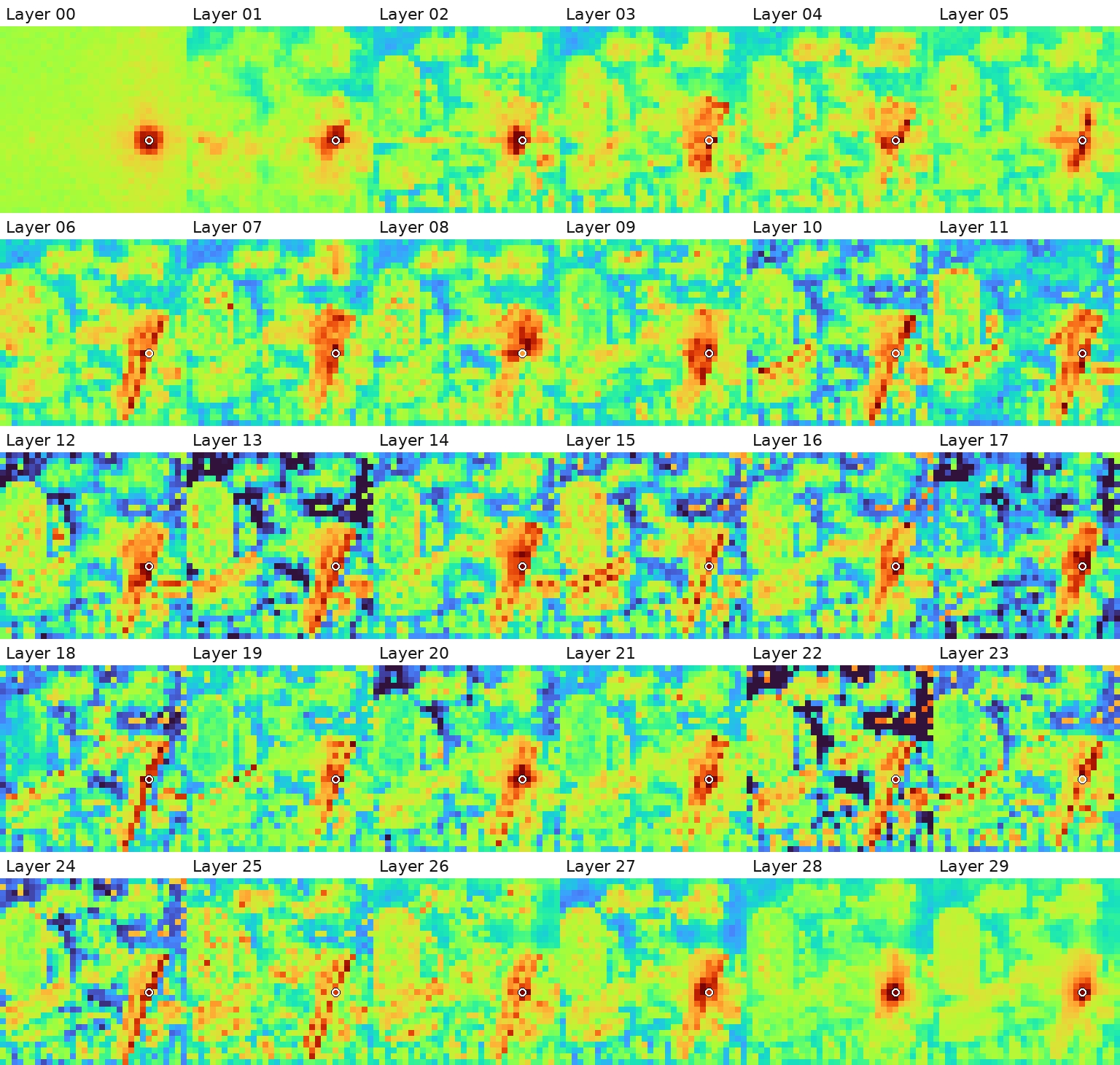}} &
    \raisebox{-0.5\height}{\includegraphics[height=0.226\textwidth]{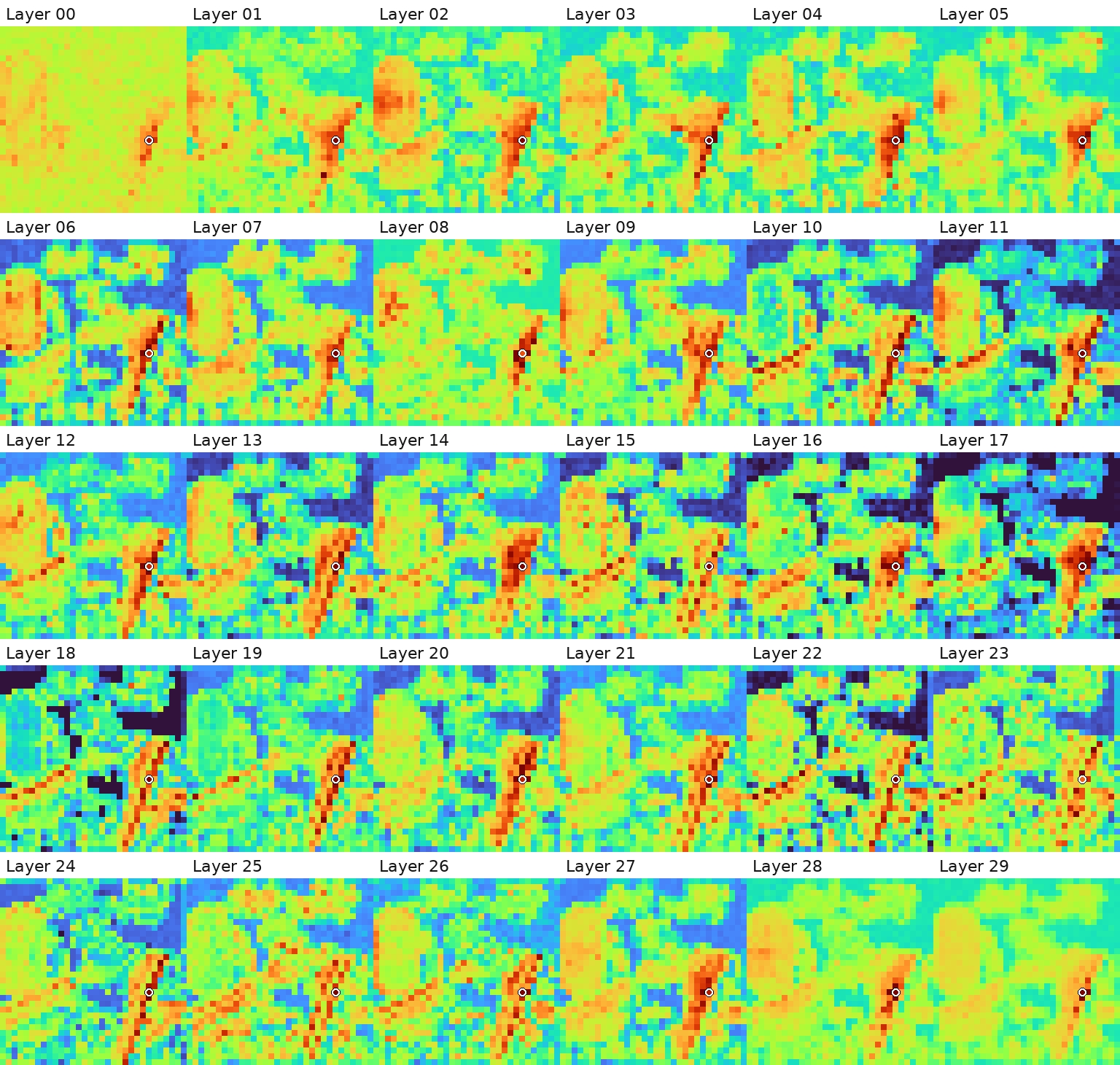}} \\
    & World space distance & Wan~2.1 RoPE & Wan~2.1 RoPE + $G_{\mathrm{buf}}$ & 3D-aware RoPE + $G_{\mathrm{buf}}$\\
    \end{tabular}
    \caption{\textbf{Attention visualization.} We show the attention activations for all layers for two query points on the lion king DTC example. 
    The RoPE of Wan~2.1 combines pixel $xy$-coordinates, $\mathbf{p}_{uv}$, and the frame id $f_{\mathrm{id}}$. We add a G-buffer wpos and normals, $G_{\mathrm{buf}}$, as conditions (middle column). In the rightmost column, we show our 3D-aware RoPE + $G_{\mathrm{buf}}$.
    Please zoom on the PDF image to see details. For reference, the left column shows the world space distance between two points in texture space. Note that our embedding attends strongly to points closer in world space over all layers of the network. 
    }
    \label{fig:rope_attn_supplemental}    
\end{figure*}
}

\newcommand{\figTEXGen}{
\begin{figure}
    \centering
    \setlength{\tabcolsep}{1pt}
    \begin{tabular}{*{6}{>{\centering\arraybackslash}p{0.16\columnwidth}}}
       \multicolumn{3}{c}{\includegraphics[width=0.49\columnwidth]{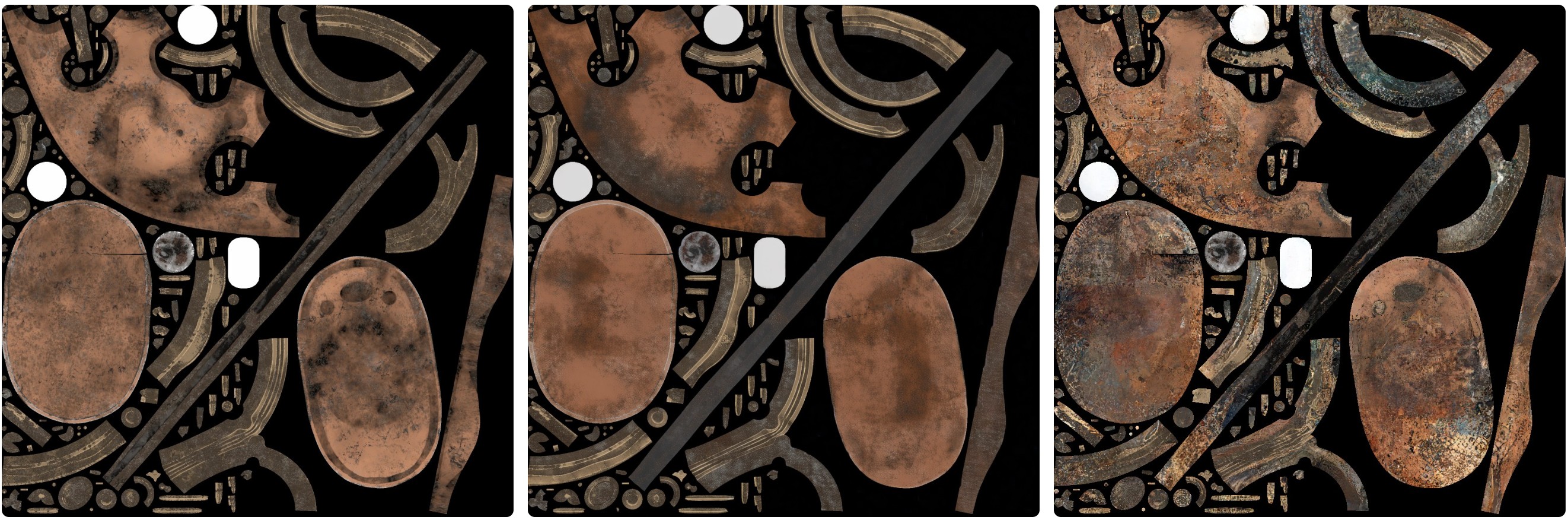}} &
       \multicolumn{3}{c}{\includegraphics[width=0.49\columnwidth]{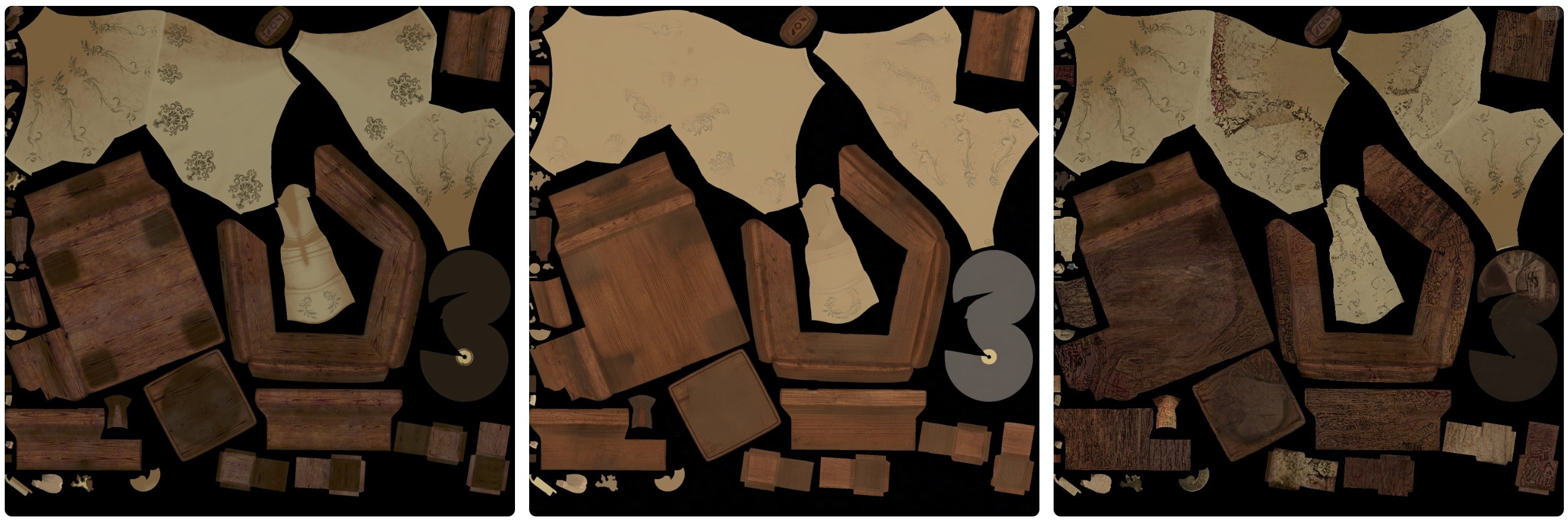}} \\
       \multicolumn{3}{c}{\includegraphics[width=0.49\columnwidth]{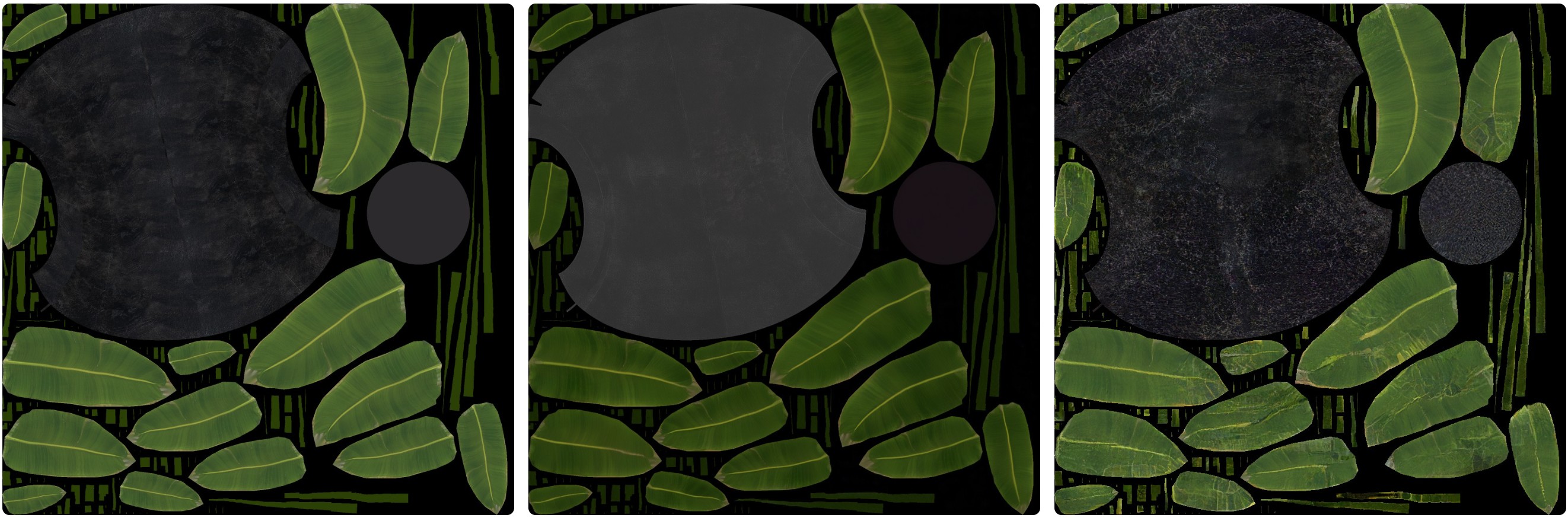}} &
       \multicolumn{3}{c}{\includegraphics[width=0.49\columnwidth]{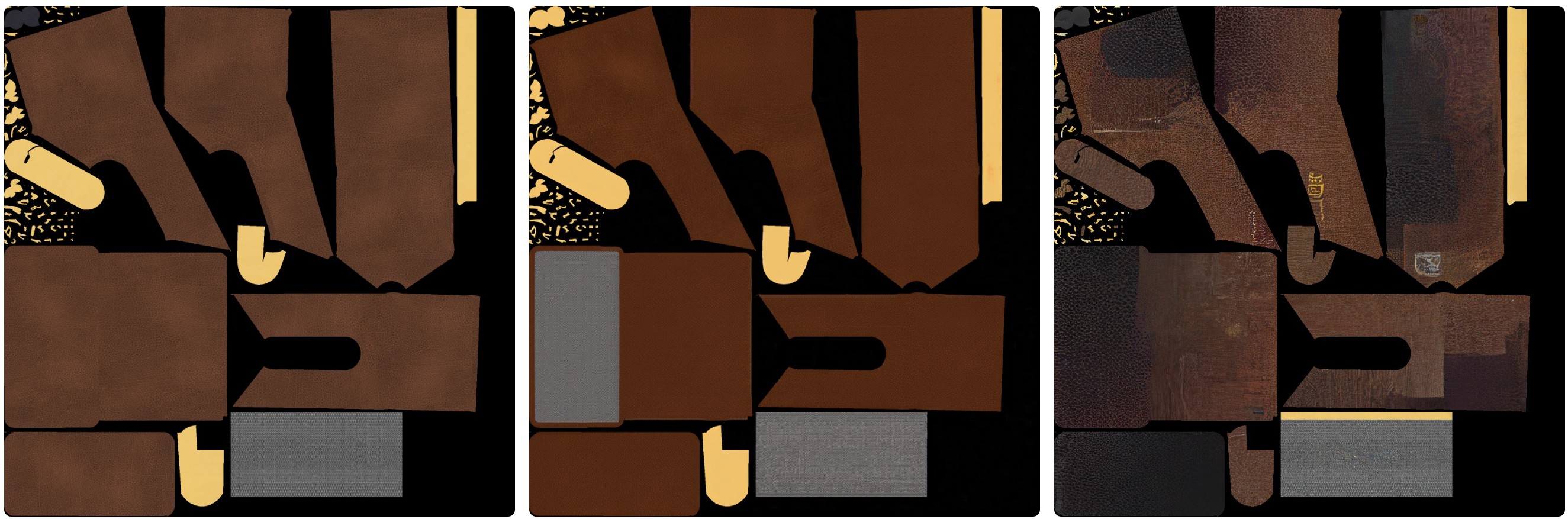}} \\
       Ref & Our & TEXGen & Ref & Our & TEXGen
    \end{tabular}
    \caption{Generated basecolor textures from single-view guidance. Overall, our method generates more consistent albedo maps than TEXGen, demodulates lighting from the conditional view (TEXGen uses a demodulated conditional view), and outputs full PBR materials (not shown here).}
    \label{fig:texgen}
\end{figure}
}

\newcolumntype{Y}{>{\centering\arraybackslash}X}
\newcolumntype{L}{>{\raggedright\arraybackslash}X}

\newcommand{\figDTC}{
\begin{figure}
    \centering
    \includegraphics[width=\textwidth]{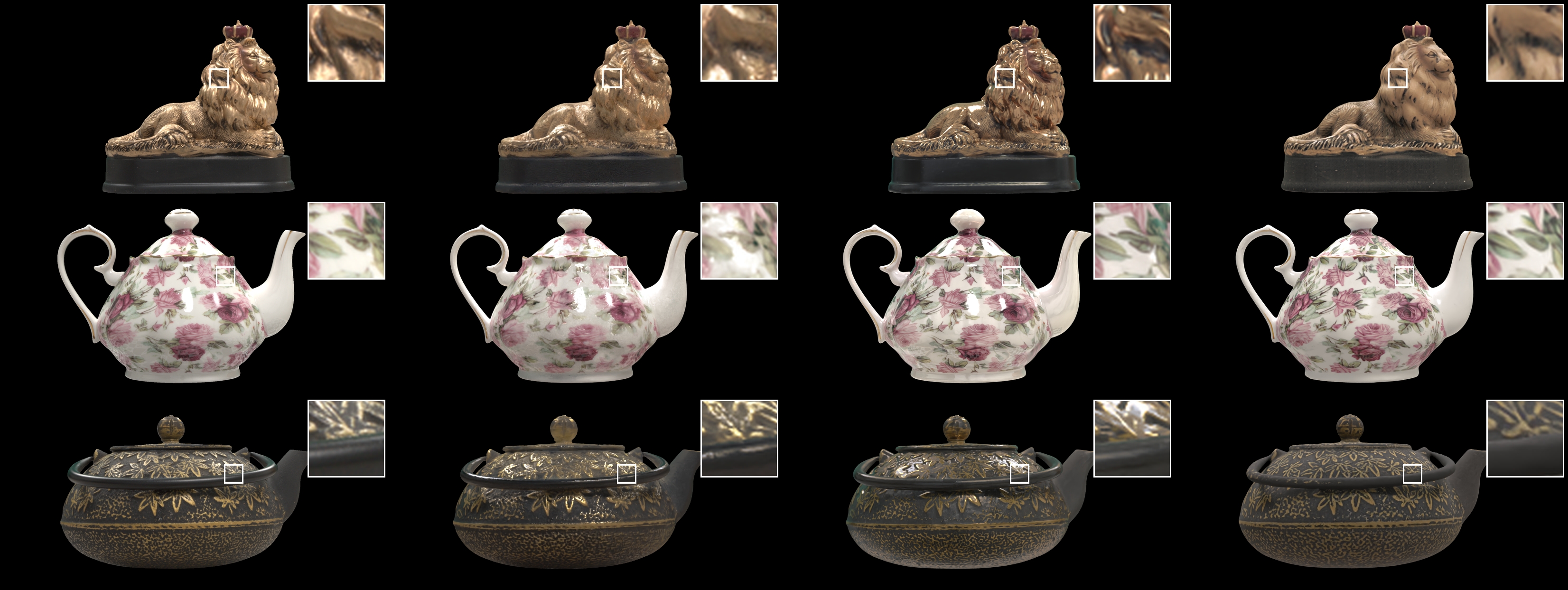}
    \setlength{\tabcolsep}{1pt}
    \begin{tabularx}{\textwidth}{YYYY}
    Input view (photo) & DiffPT & Our & LSRM \\
    \end{tabularx}
    \caption{\textbf{Reconstruction~-~Real-World.} Material reconstructions from multi-view observations (photographs with known poses) on three examples from the DTC dataset~\citep{dong2025dtc}. 
    For optimization-based differentiable path tracing (DiffPT) and our method, we leverage known geometry, while LSRM jointly reconstructs geometry and materials.}
    \label{fig:dtc_scenes}
\end{figure}
}

\newsavebox{\dtcimagebox}
\newlength{\dtcimagewidth}

\newcommand{\dtcrowlabel}[1]{%
    \vbox to \dimexpr\ht\dtcimagebox/3\relax{%
        \vfil
        \hbox to 1.6em{%
            \hfil
            \rotatebox[origin=c]{90}{#1}%
            \hfil
        }%
        \vfil
    }%
}

\newcommand{\figDTCmain}{%
\begin{figure}
    \centering
    \small

    \setlength{\dtcimagewidth}{0.95\linewidth}

    \sbox{\dtcimagebox}{%
        \includegraphics[
            width=\dtcimagewidth
        ]{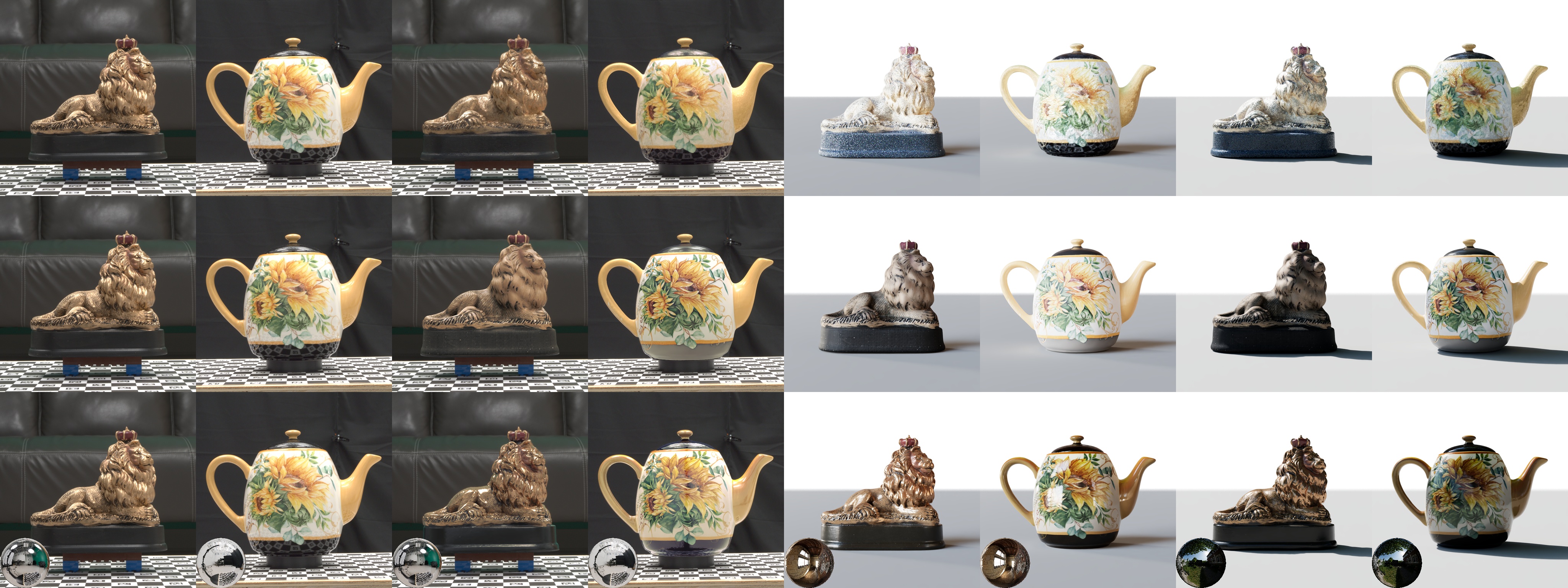}%
    }

    \setlength{\tabcolsep}{0pt}

    \begin{tabular}{@{}c@{}c@{}}
        \makebox[0pt][r]{%
            \vbox{%
                \offinterlineskip
                \dtcrowlabel{DiffPT}%
                \dtcrowlabel{LSRM}%
                \dtcrowlabel{Ours}%
            }%
            \hspace{-2pt}%
        }
        &
        \usebox{\dtcimagebox}
        \\[3pt]

        {}
        &
        \begin{tabularx}{\dtcimagewidth}{
            @{}
            *{4}{>{\centering\arraybackslash}X}
            @{}
        }
            Input view (photo) &
            Reconstructed &
            Relit 1 &
            Relit 2
        \end{tabularx}
    \end{tabular}

    \caption{%
        \textbf{Reconstruction~-~Real-World.}
        Material reconstructions from multi-view observations
        (photographs with known poses) on two examples from the DTC
        dataset~\citep{dong2025dtc}. For optimization-based differentiable path tracing (DiffPT)
        and our method, we leverage known geometry, while LSRM jointly reconstructs
        geometry and materials. In the right part we show the
        reconstructed materials under two novel lighting conditions.
    }
    \label{fig:dtc}
\end{figure}%
}

\newcommand{\figVACEsmall}{
\begin{figure}
    \centering

    \begin{minipage}[t]{0.60\textwidth}
        \vspace{0pt}
        \centering
        \small
        \setlength{\tabcolsep}{1pt}
    
        \begin{tabularx}{\linewidth}{@{}YYYY@{}}
            Reference & Ours & Ours & DiffPT \\
            & {\tiny single view} & {\tiny VACE} & {\tiny VACE} \\
            \multicolumn{4}{@{}c@{}}{%
                \includegraphics[width=\linewidth]{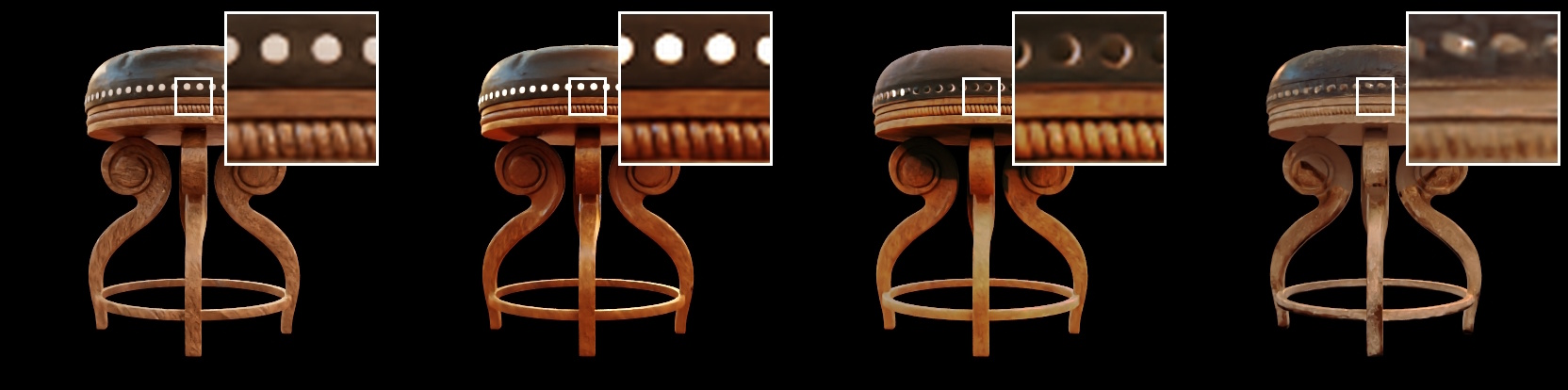}%
            }
        \end{tabularx}
    \end{minipage}
    \hfill
    \begin{minipage}[t]{0.39\textwidth}
        \vspace{0pt}
        \centering
        \small
        \setlength{\tabcolsep}{1pt}
        \begin{tabularx}{\linewidth}{@{}LYYY@{}}
            Metrics & Ours & Ours & DiffPT \\
           {\tiny 1088 views} & {\tiny single view} & {\tiny VACE} & {\tiny VACE} \\
            \toprule
            {\tiny PSNR$\uparrow$}       & 29.81  & 27.07  & 23.37  \\
            {\tiny SSIM$\uparrow$}       & 0.960  & 0.946  & 0.907  \\
            {\tiny LPIPS$\downarrow$}    & 0.0364 & 0.0509 & 0.0762 \\
            {\tiny CLIP-FID$\downarrow$} & 1.97   & 3.06   & 5.29   \\
            {\tiny CMMD$\downarrow$}     & 0.027  & 0.036  & 0.132  \\
            \bottomrule
        \end{tabularx}        
    \end{minipage}
    \vspace*{-3mm}
    \caption{\textbf{Reconstruction~-~Generated Frames.} Material reconstructions from imperfect views, with views synthesized by an off-the-shelf
    depth-conditioned video model: Wan2.2-VACE-Fun-A14B~\citep{vace2025}. Compared to an optimization-based approach, 
    our method robustly reconstructs materials also from the inconsistent views from the video model.}
    \label{fig:vace}
\end{figure}
}

\newcommand{\figTEXGENmain}{
\begin{figure}
    \centering

    \begin{minipage}[t]{0.65\textwidth}
        \vspace{0pt}
        \centering
        \small
        \setlength{\tabcolsep}{1pt}
        \begin{tabularx}{\linewidth}{@{}Y@{}Y@{}Y@{}}
            TEXGen & Ours & Reference \\
            \includegraphics[width=0.9\linewidth]{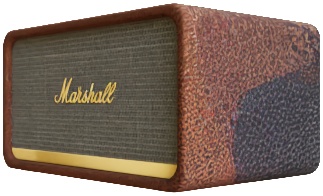} & 
            \includegraphics[width=0.9\linewidth]{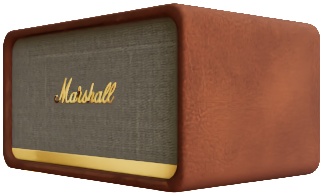} & 
            \includegraphics[width=0.9\linewidth]{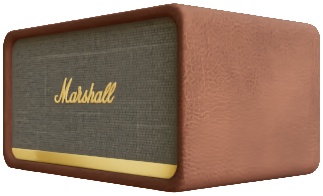} \\
        \end{tabularx}
    \end{minipage}
    \hfill
    \begin{minipage}[t]{0.34\textwidth}
        \vspace{0pt}
        \centering
        \small
        \setlength{\tabcolsep}{1pt}
        \begin{tabularx}{\linewidth}{@{}LYY@{}}
            Metrics & TEXGen & Ours \\
            \toprule
            CLIP-FID$\downarrow$ & 3.790  & \sota{1.998}    \\
            CMMD$\downarrow$     & 0.0867 & \sota{0.0067}   \\
            LPIPS$\downarrow$    & 0.0459 & \sota{0.0305}  \\
            \bottomrule
        \end{tabularx}        
    \end{minipage}
    
    \caption{
    \textbf{TEXGen}. TEXGen~\citep{yu2024texgen} only produces albedo textures. Metrics for \textbf{diffuse-only} renderings from 32 scenes $\times$ 8 views $\times$ 8 probes from single-view guidance.}
    \label{fig:texgen_main}
\end{figure}
}

\newcommand{\figInfAblation}{%
\begin{figure}
    \centering
    \setlength{\tabcolsep}{0pt}

    \begin{tabularx}{0.98\textwidth}{@{}YYY@{}}
        \multicolumn{3}{c}{\includegraphics[width=0.98\textwidth]{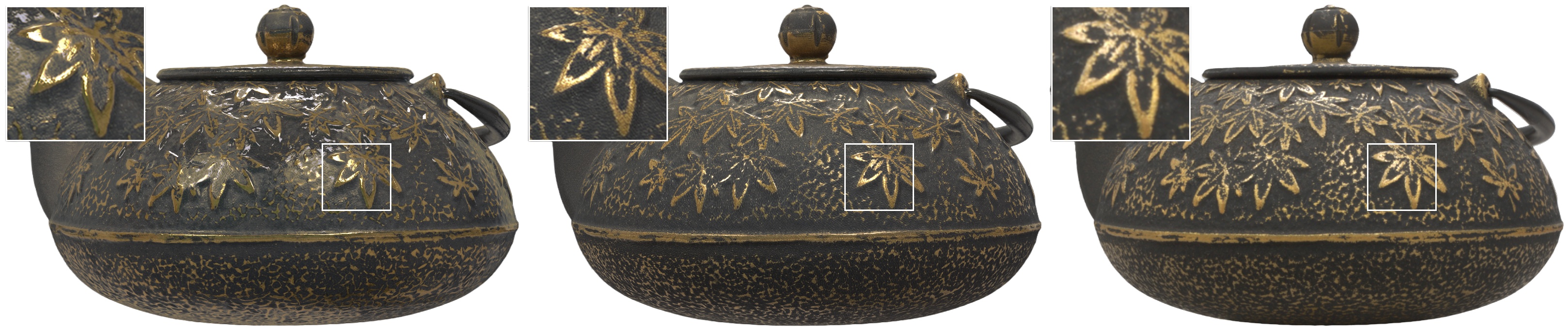}}
        \\
        17 views ~--~ 2K & 117 views  ~--~ 8K & Reference \\
        \sota{28.42dB} / 0.919 / 0.0668 & 28.34dB / \sota{0.917} / \sota{0.0639} & PSNR~$\uparrow$ / SSIM~$\uparrow$ / LPIPS~$\downarrow$
    \end{tabularx}

    \caption{\textbf{Ablation~--~Inference Scaling.}
    With our inference augmentations, we can scale our results to 8K
    texture resolution and increase the number of input views from 17
    to 117.
    This enables capturing high-frequency details and accurate specular predictions.
    We provide average metrics over the first four samples of the real dataset, showing preserved fidelity with slightly improved LPIPS.    
    }
    \label{fig:teapot}
\end{figure}%
}

\newcommand{\figSingleView}{
\begin{figure}
    \centering
    \includegraphics[width=\textwidth]{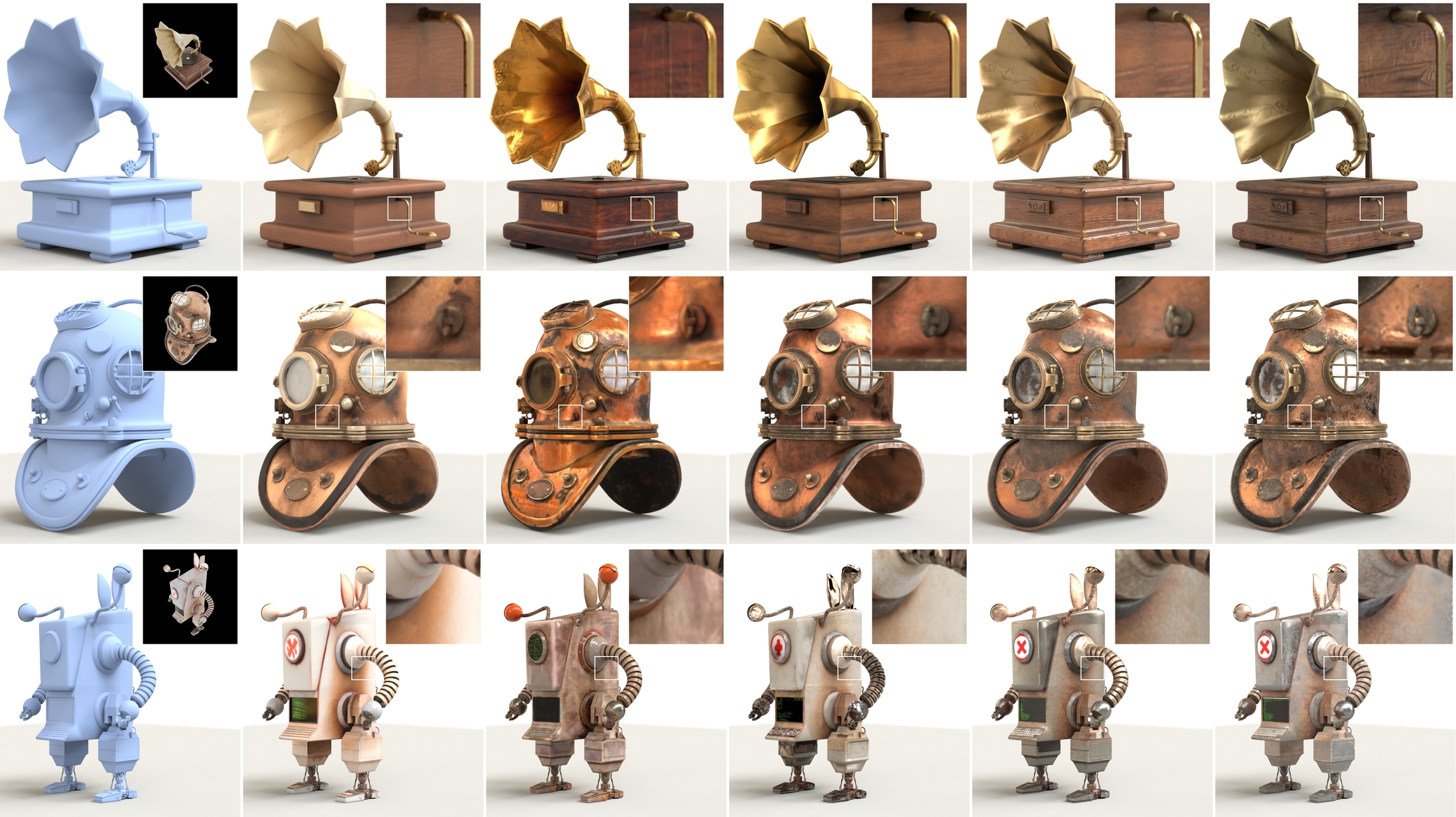}
    \setlength{\tabcolsep}{1pt}
    \begin{small}
    \begin{tabularx}{\textwidth}{YYYYYY}
    Geo & Hunyuan3D~2.1 & VideoMatGen & Trellis.2 & \textbf{Ours} & Reference
    \end{tabularx}
    \end{small}
    \caption{
    \textbf{Generation~-~Single-View.} Materials from our generative method, conditioned on a single image and text description. The visualized view is rotated slightly from the conditioning view (leftmost insets). We use reference geometry and the same conditioning view for all methods.
    }
    \label{fig:single_view}
\end{figure}
}

\newcommand{\figTextToMaterial}{
\begin{figure}[b]
    \centering
    \includegraphics[width=\textwidth]{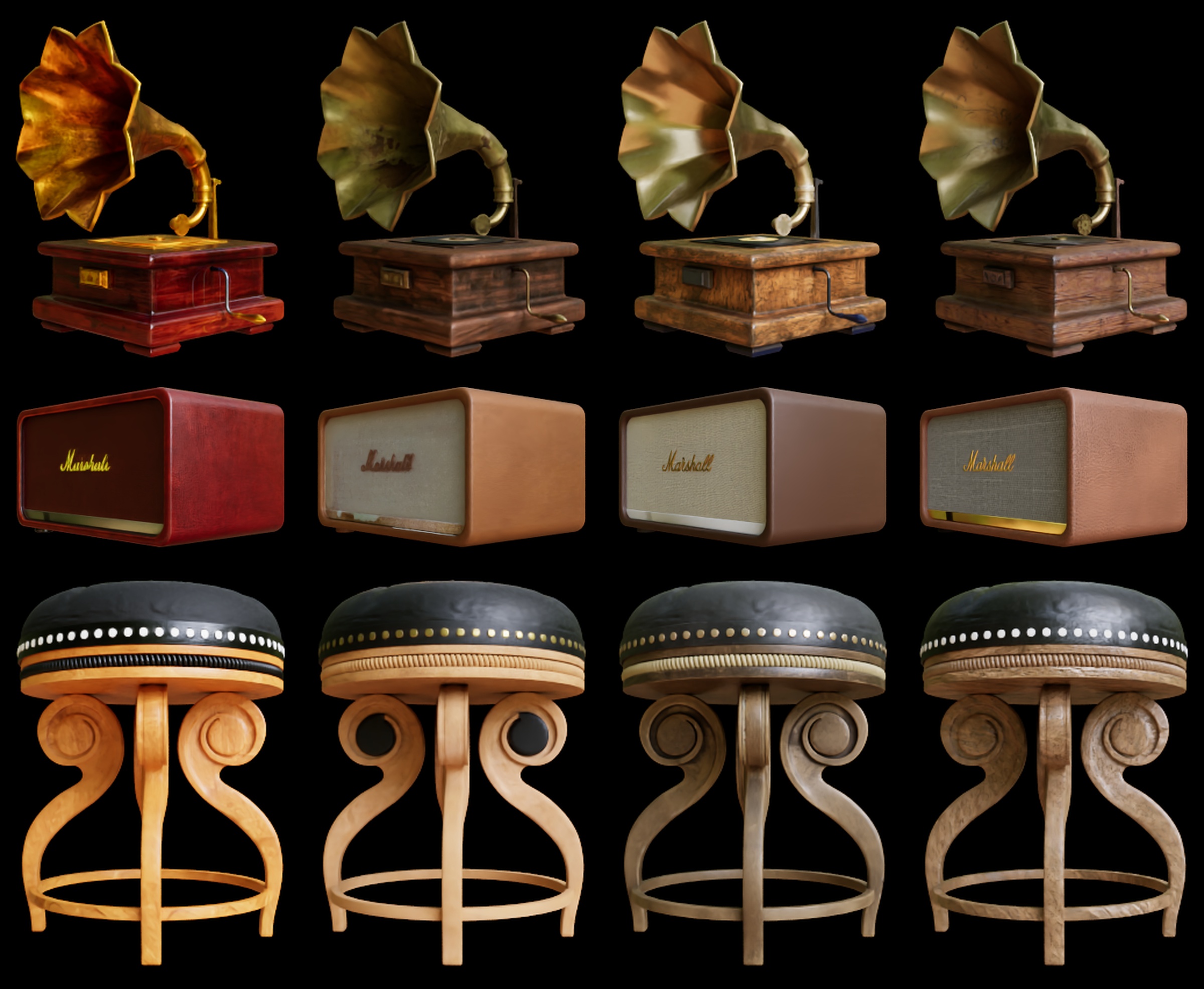}
    \setlength{\tabcolsep}{1pt}
    \begin{small}
    \begin{tabularx}{\textwidth}{YYYY}
    VideoMatGen & VideoMat & Our & Reference\\
    \end{tabularx}
    \end{small}
    \caption{
    \textbf{Generation~-~Text to material.} Materials from our generative method, conditioned on a text prompt and the input geometry (the UV mask, world space positions, and surface normals). We compare against VideoMat~\citep{munkberg2025videomat} and VideoMatGen~\citep{hasselgren2026videomatgen}. Without any image guidance, our model still produces semantically meaningful materials for all examples.
    }
    \label{fig:text_to_material}
\end{figure}
}

\newcolumntype{R}{>{\centering\arraybackslash}m{1.5em}}

\newcommand{\componentwidth}{0.24}

\newcommand{\figComponents}{
\begin{figure}
\centering
\setlength{\tabcolsep}{1pt}
\small
\begin{tabular}{cc@{}c@{}c@{}c@{}}
\rotatebox[origin=c]{90}{Base Color (Adjusted)} &
\raisebox{-0.5\height}{\includegraphics[width=\componentwidth\columnwidth]{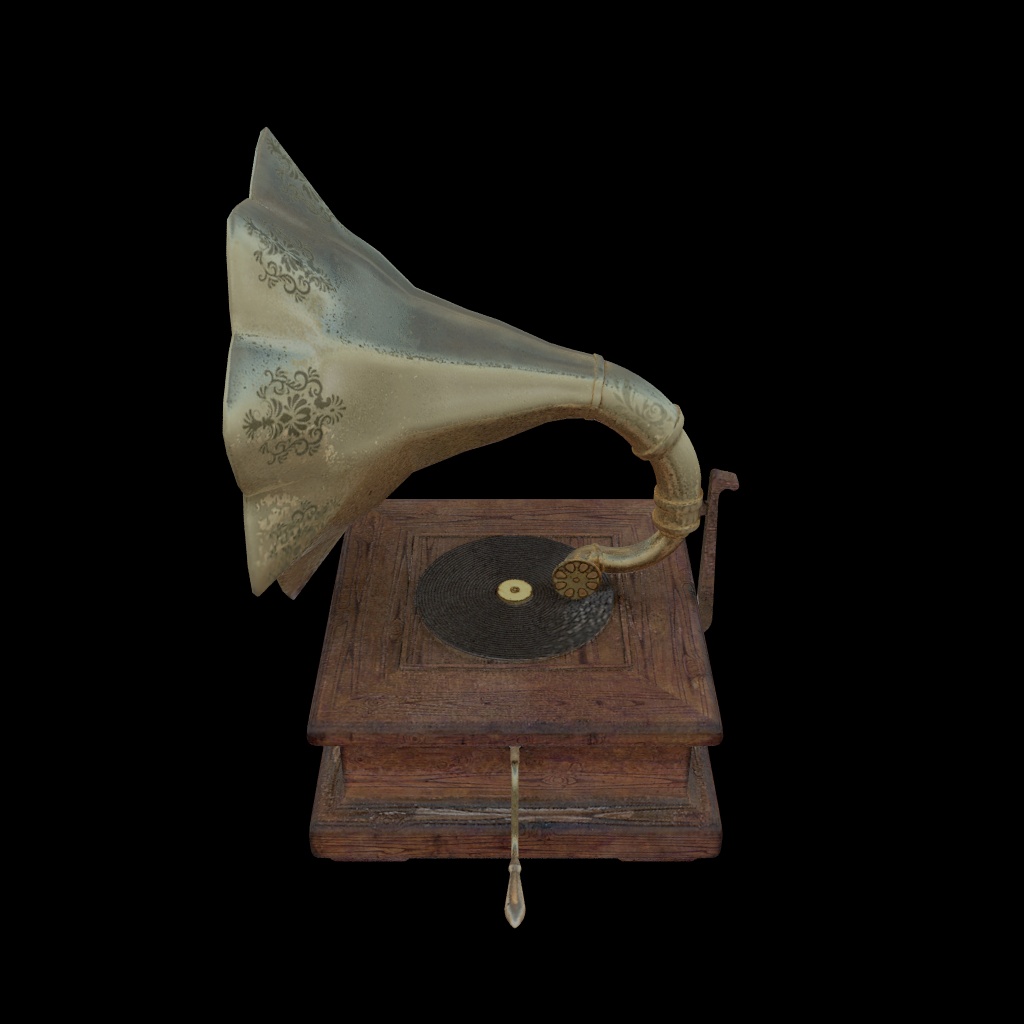}} &
\raisebox{-0.5\height}{\includegraphics[width=\componentwidth\columnwidth]{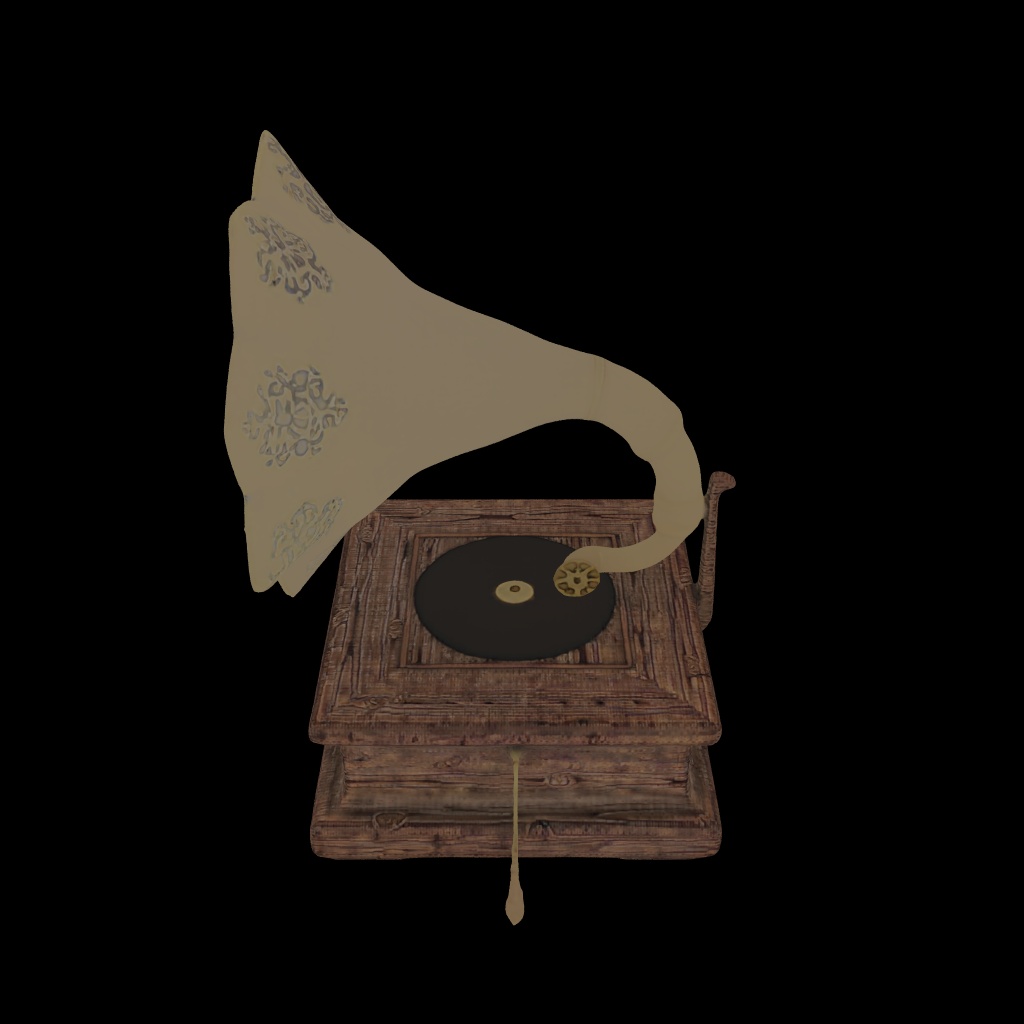}} &
\raisebox{-0.5\height}{\includegraphics[width=\componentwidth\columnwidth]{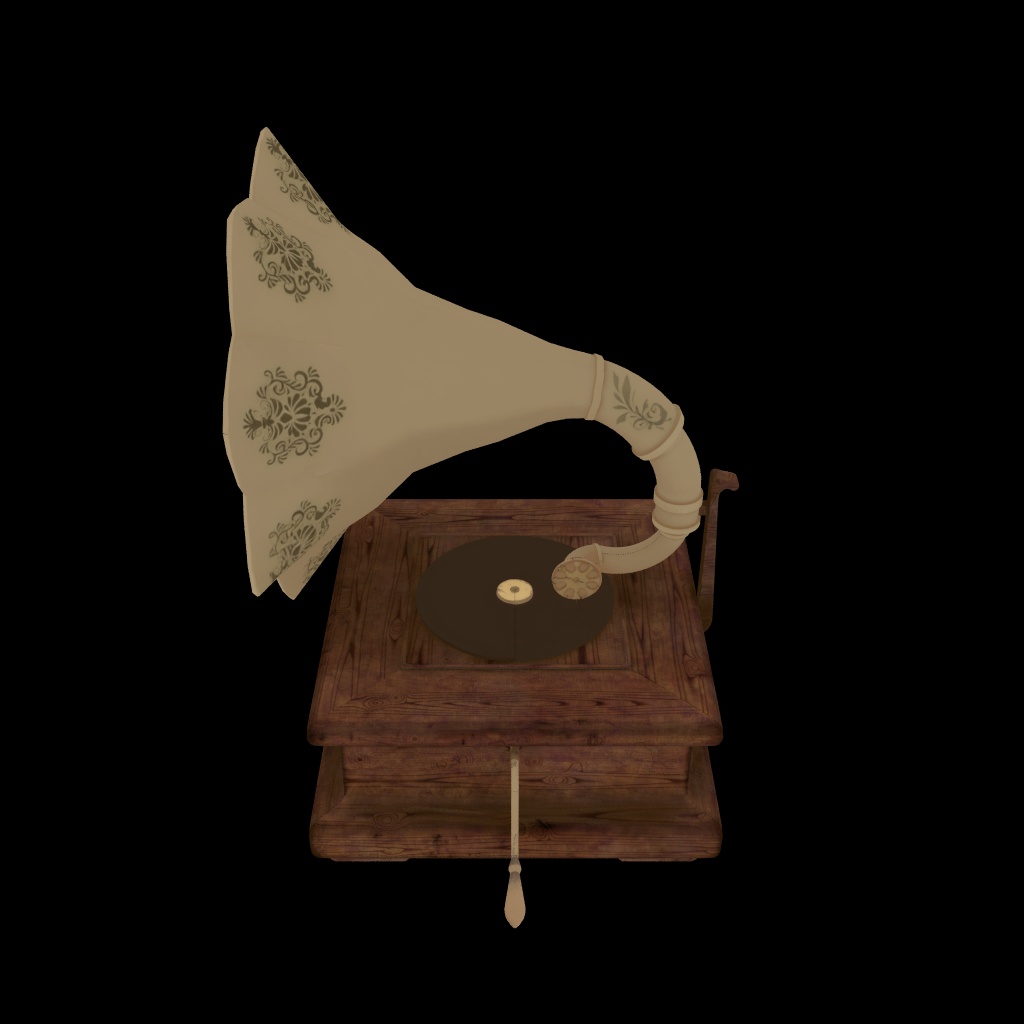}} &
\raisebox{-0.5\height}{\includegraphics[width=\componentwidth\columnwidth]{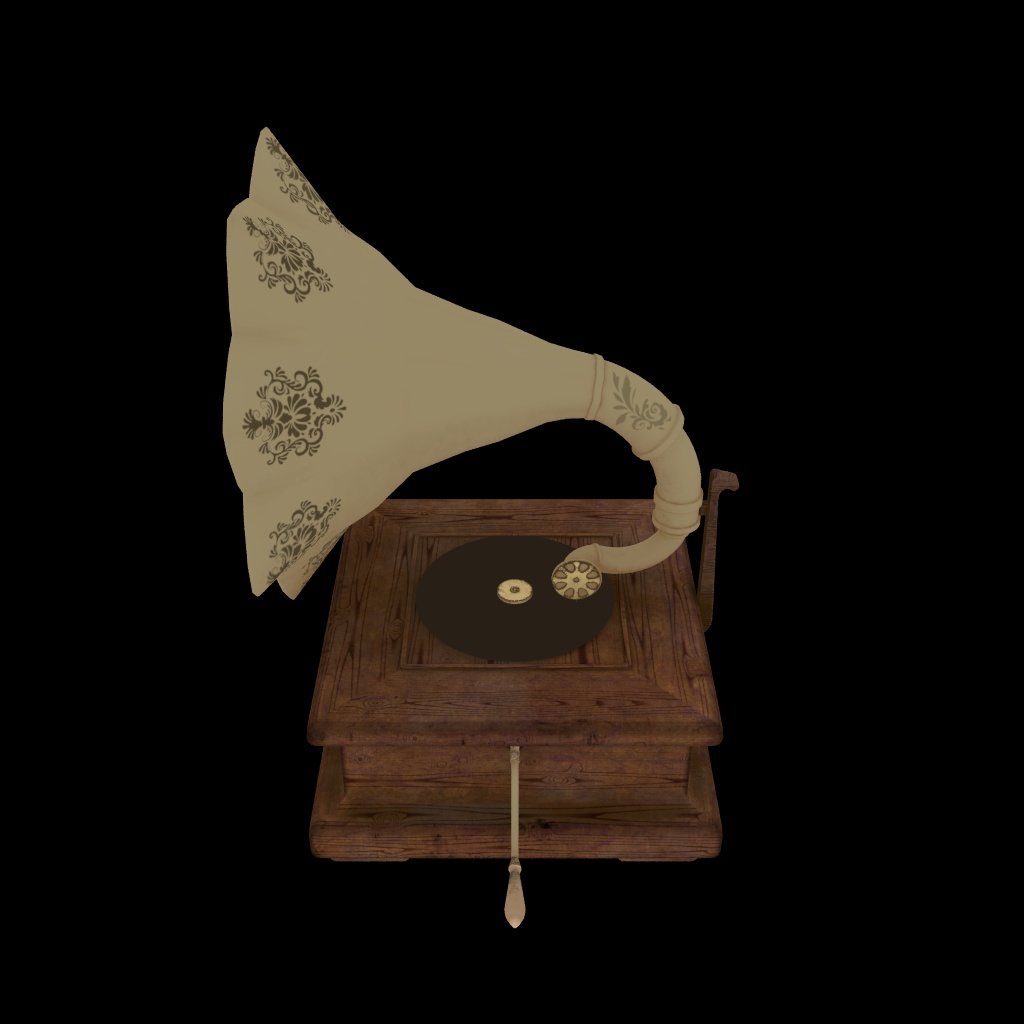}}
\\

\rotatebox[origin=c]{90}{Roughness} &
\raisebox{-0.5\height}{\includegraphics[width=\componentwidth\columnwidth]{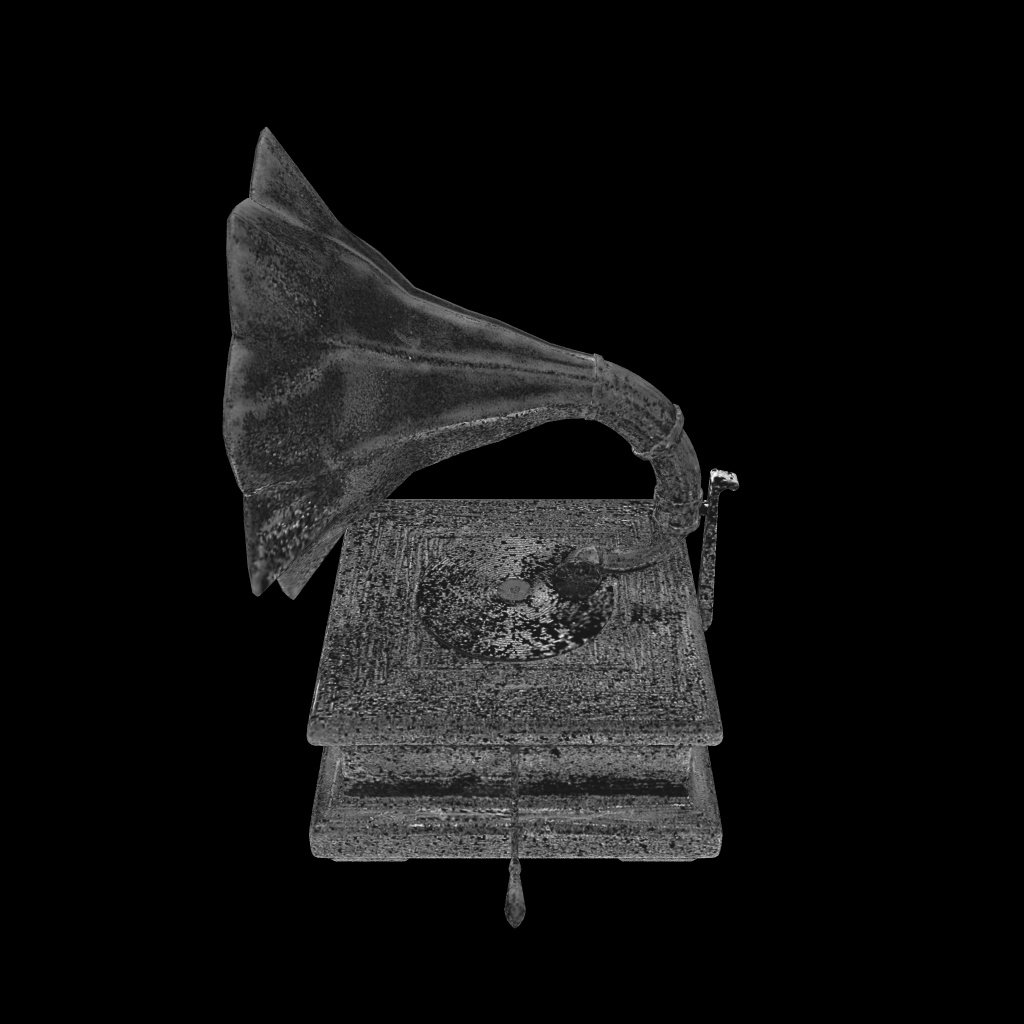}} &
\raisebox{-0.5\height}{\includegraphics[width=\componentwidth\columnwidth]{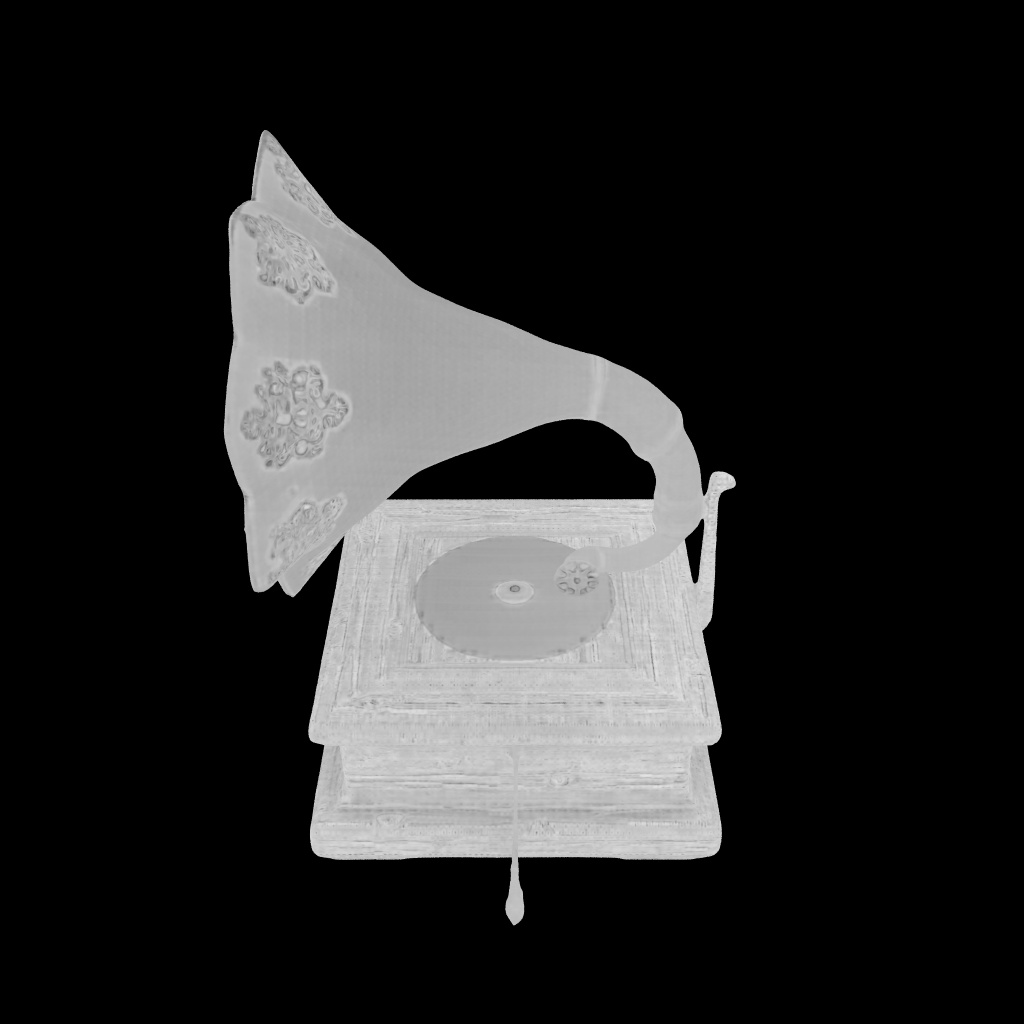}} &
\raisebox{-0.5\height}{\includegraphics[width=\componentwidth\columnwidth]{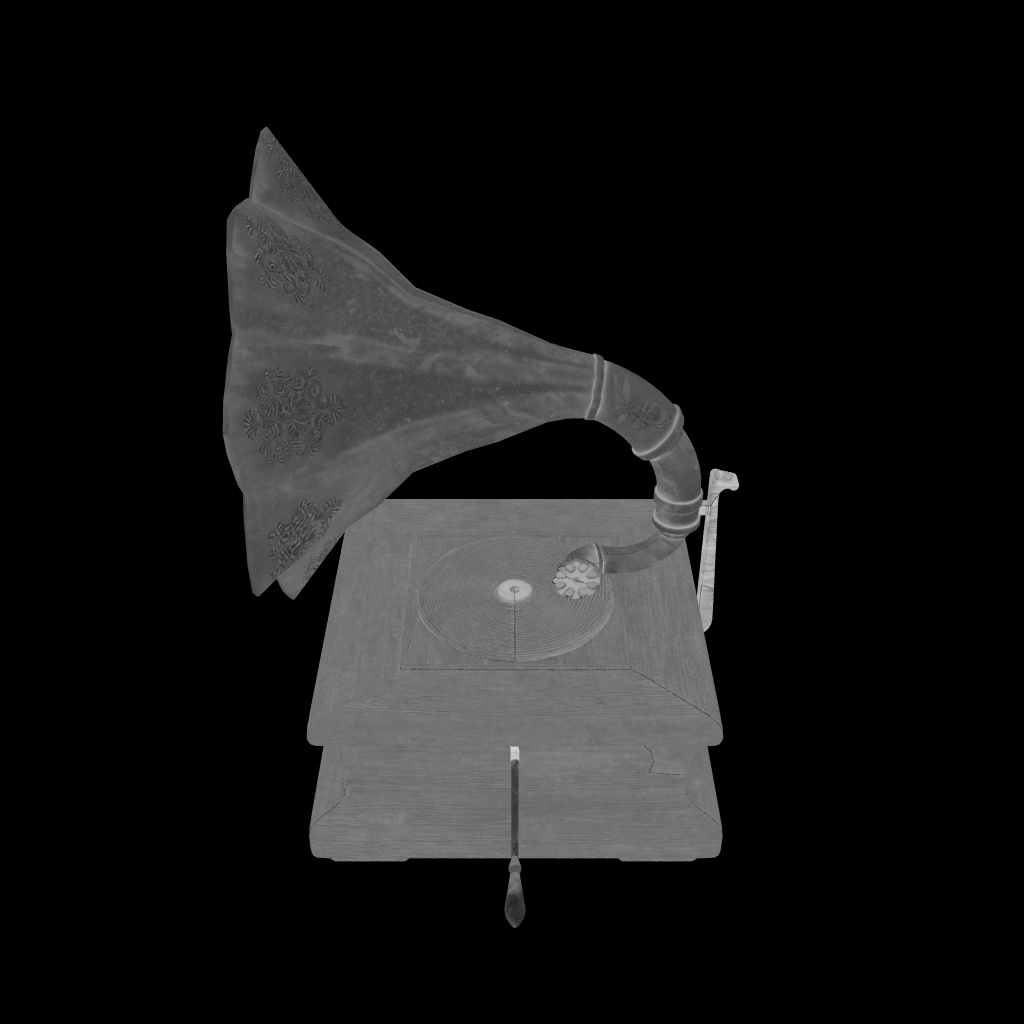}} &
\raisebox{-0.5\height}{\includegraphics[width=\componentwidth\columnwidth]{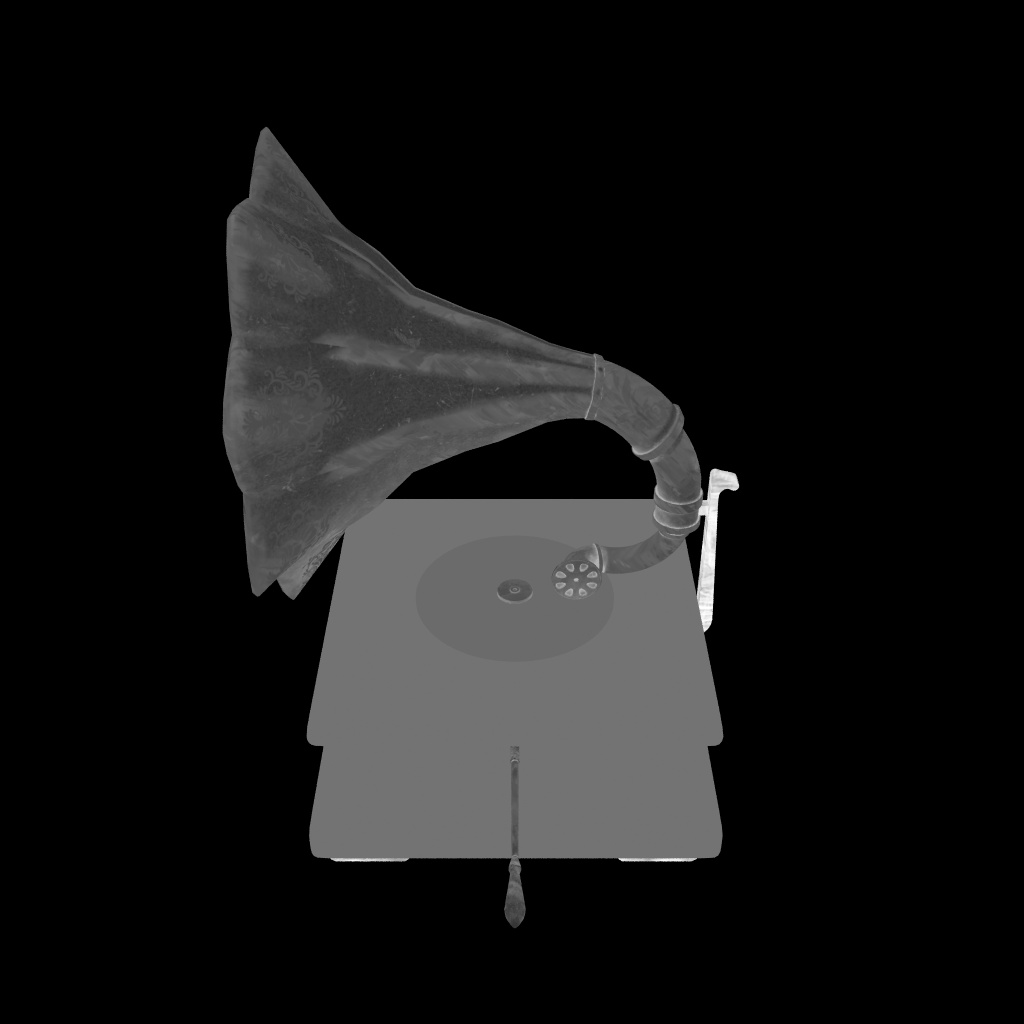}}
\\

\rotatebox[origin=c]{90}{Metallic} &
\raisebox{-0.5\height}{\includegraphics[width=\componentwidth\columnwidth]{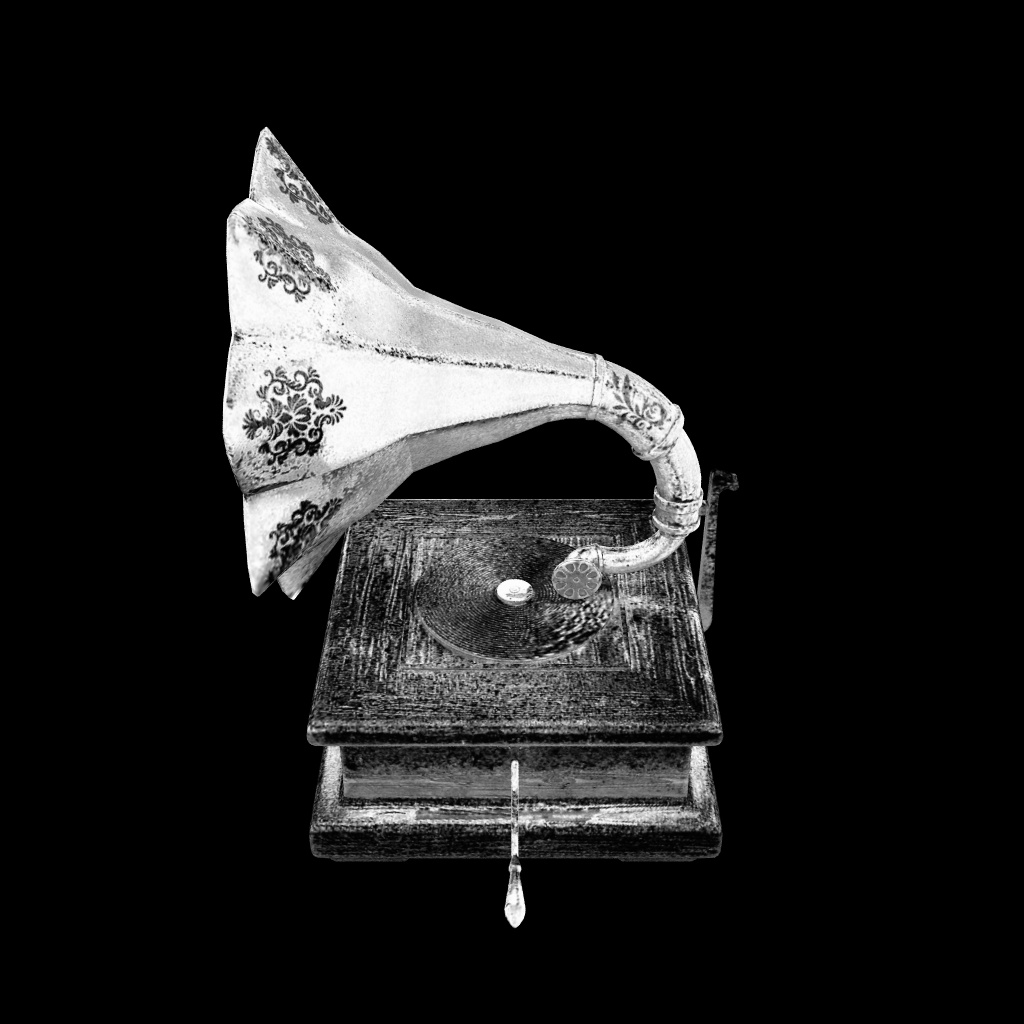}} &
\raisebox{-0.5\height}{\includegraphics[width=\componentwidth\columnwidth]{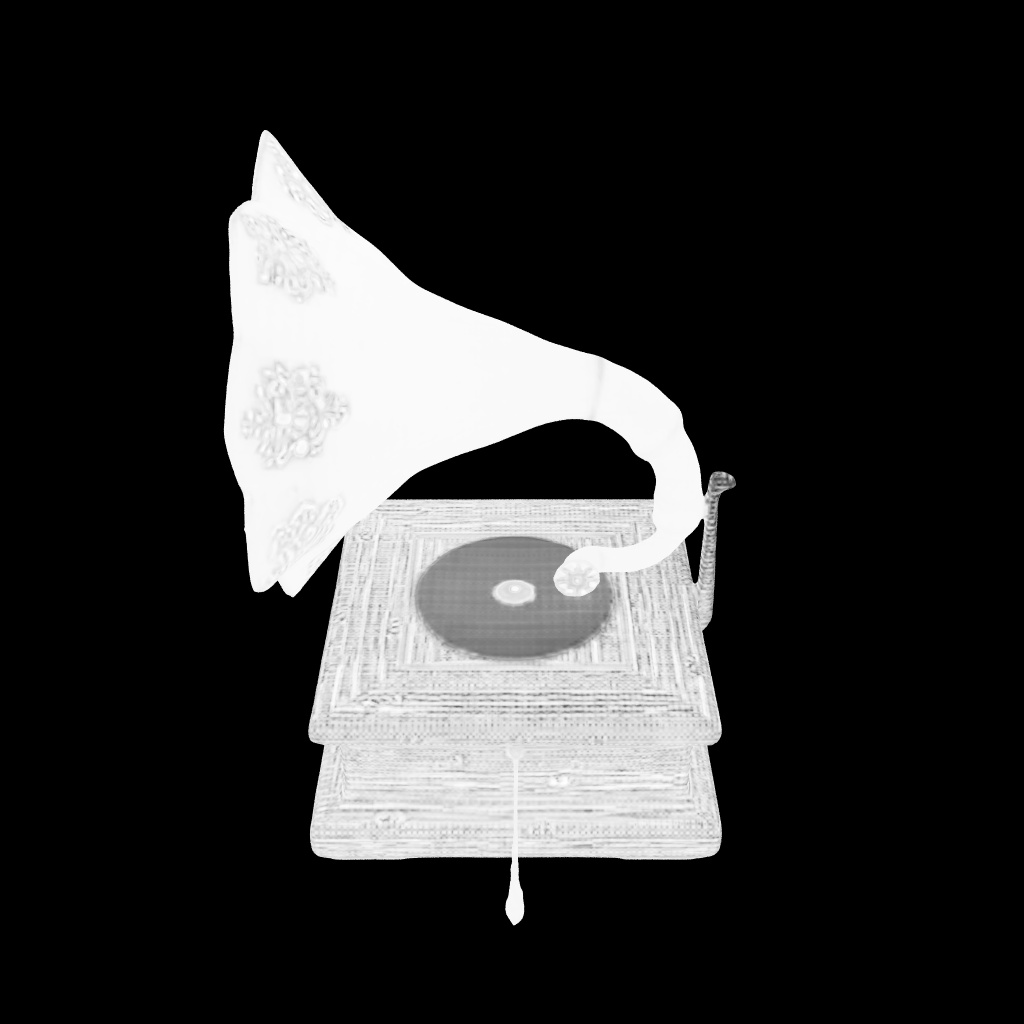}} &
\raisebox{-0.5\height}{\includegraphics[width=\componentwidth\columnwidth]{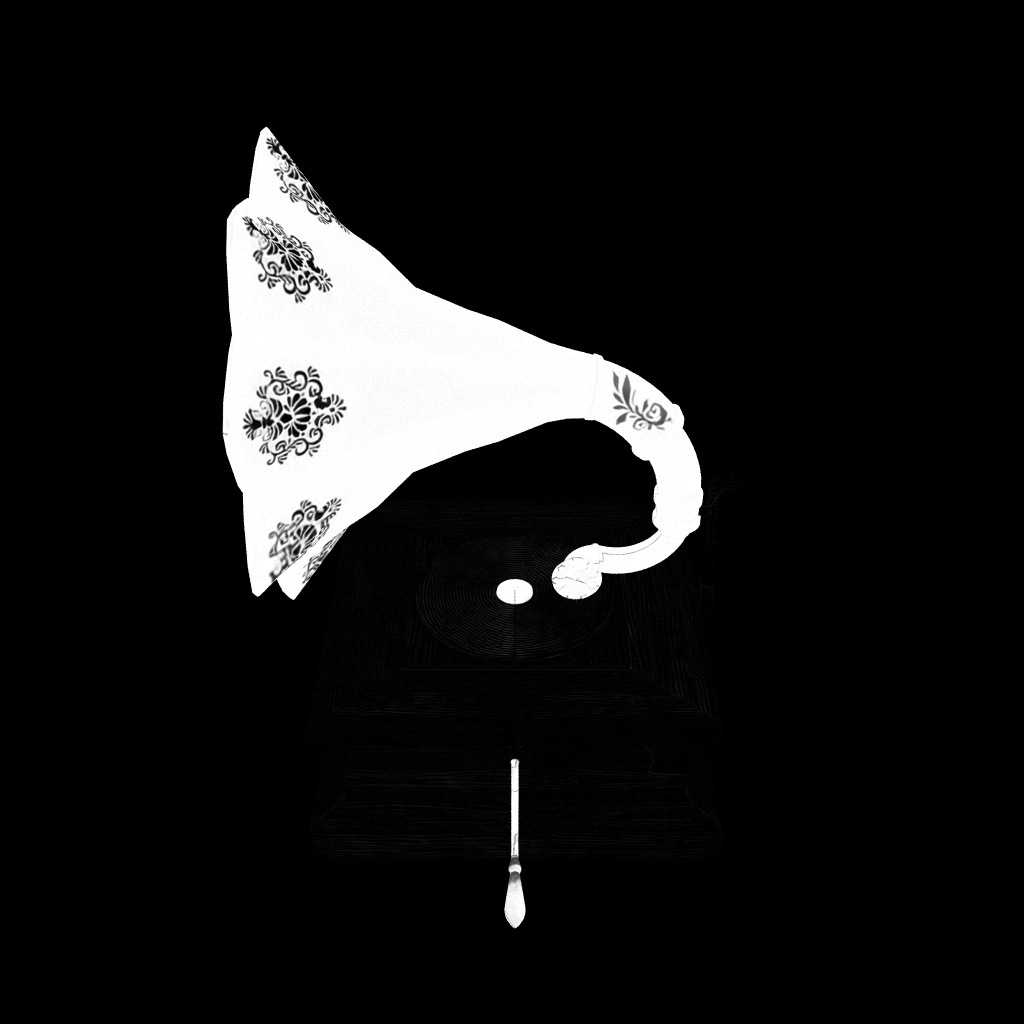}} &
\raisebox{-0.5\height}{\includegraphics[width=\componentwidth\columnwidth]{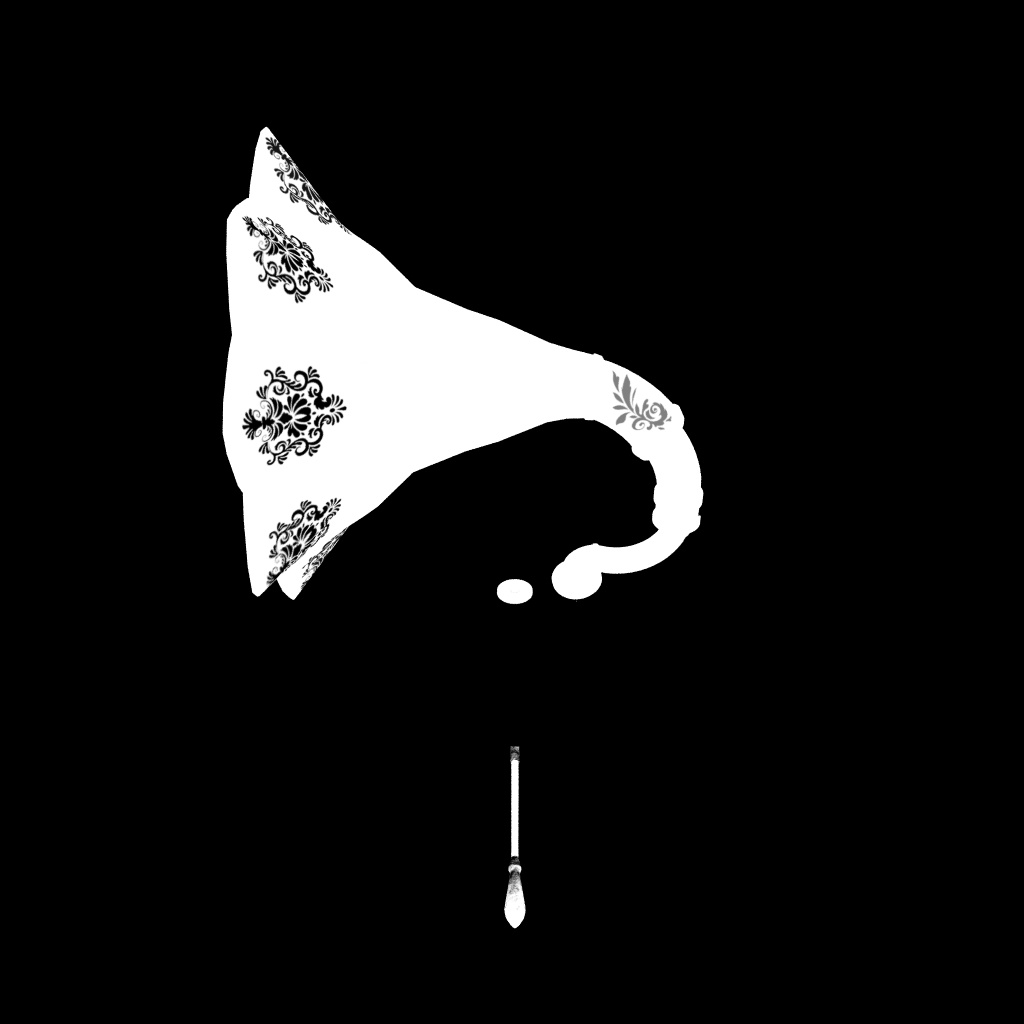}}
\\

& DiffPT & LSRM & Ours & Reference
\end{tabular}

\caption{\textbf{Material components}.}
\label{fig:mat_component}
\end{figure}
}

\newcommand{\figCRF}{
\begin{figure}
\centering
\setlength{\tabcolsep}{2pt}
\begin{tabular}{c c c}
\includegraphics[width=0.32\columnwidth]{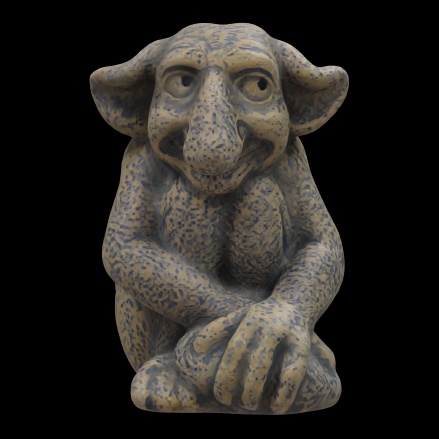} &
\includegraphics[width=0.32\columnwidth]{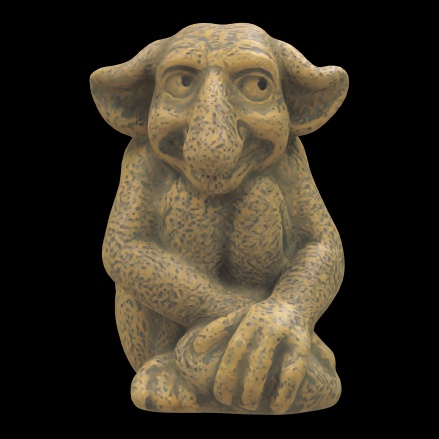} &
\includegraphics[width=0.32\columnwidth]{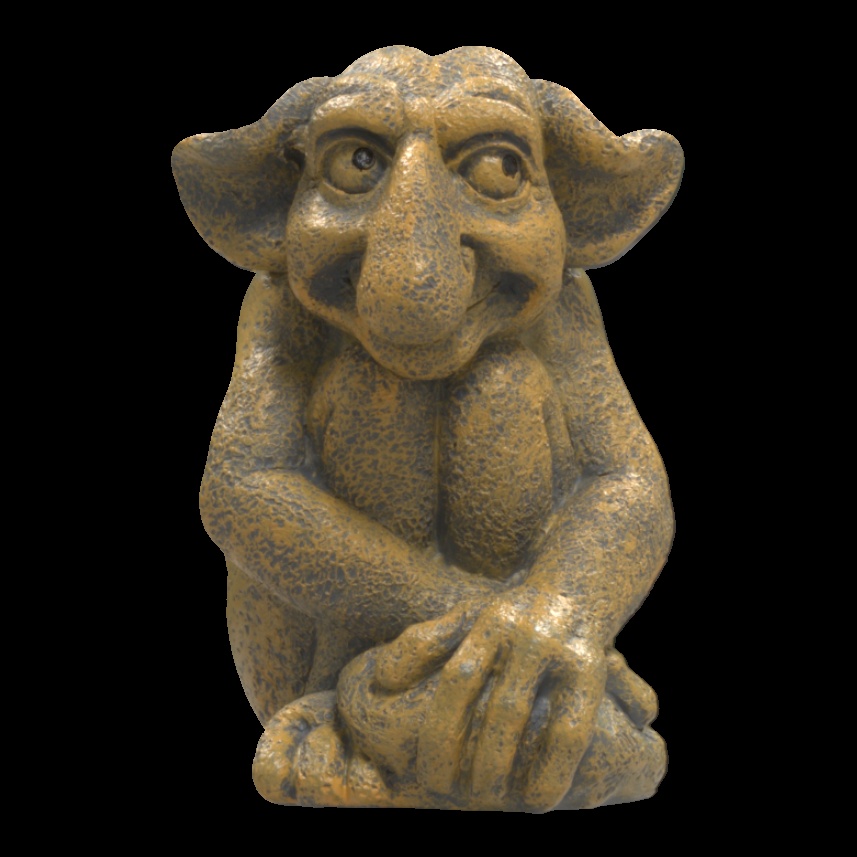} \\

w/o CRF & w/ CRF & Reference
\end{tabular}
\caption{\textbf{Effect of CRF}.
    We show the effect of the CRF adjustment on the LSRM~\citep{lsrm2026} prediction of the Gargoyle sample from the DTC dataset ~\citep{dong2025dtc}. 
    Even though the material patterns are mostly captured properly by the baseline, color is shifted due to unknown lighting condition and potential training bias.
    CRF adjustment helps to rule out this bias and provide a more fair comparison. 
}
\label{fig:crf}
\end{figure}
}

\newcolumntype{Q}{%
  >{\hsize=1.90909\hsize\centering\arraybackslash}X%
}
\newcolumntype{N}{%
  >{\hsize=0.63636\hsize\centering\arraybackslash}X%
}

\newcommand{\figAttention}{

\begin{figure}
\centering
  \begin{tikzpicture}
    \node[anchor=south west, inner sep=0] (image) at (0,0) {%
      \includegraphics[width=\linewidth]{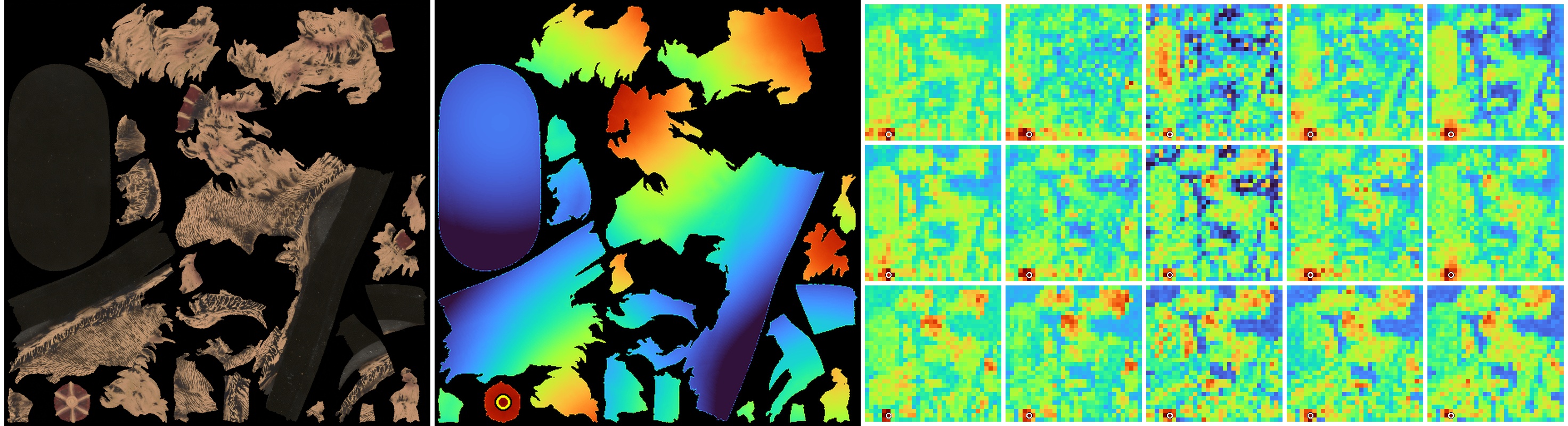}%
    };
    \draw[red, line width=1pt]
        ([xshift=18pt,yshift=6pt]image.south west)
        circle[radius=2pt];
  \end{tikzpicture}%
\\

{\small
\setlength{\tabcolsep}{0pt}%
\begin{tabularx}{\linewidth}{@{}QQNNNNN@{}}
Texture space
& World space distance
& $\ell_{3}$
& $\ell_{9}$
& $\ell_{15}$
& $\ell_{21}$
& $\ell_{27}$ \\
\end{tabularx}
}

\caption{\textbf{Attention visualization}.
 We visualize the attention activations for a query point (red circle) for three variants: \textbf{Top}: The RoPE of Wan~2.1, combining pixel $xy$-coordinates, $\mathbf{p}_{uv}$, and the frame id $f_{\mathrm{id}}$. \textbf{Middle}:
 The Wan~2.1 RoPE and G-buffer world position and normal guides, $G_{\mathrm{buf}}$. 
 \textbf{Bottom}: Our 3D-aware RoPE + $G_{\mathrm{buf}}$. 
 We show the attention activations for five layers. We note that our 3D-aware RoPE has stronger activations at similar 3D locations.}
\label{fig:attention}
\end{figure}

}

\newcommand{\tabMultiview}{

\begin{table}[tb]
\caption{\textbf{Multi-view to materials.} Comparison on synthetic materials and posed real photographs. Synthetic metrics are averaged over 32 extracted materials rendered from eight views. For the relighting metrics, we render each view with eight different HDR probes. Real metrics are averaged over eight examples from the DTC dataset~\citep{dong2025dtc}. Materials are reconstructed from 17 views in 2K resolution. We compare against
DiffPT~\citep{hasselgren2022nvdiffrecmc} and LSRM~\citep{lsrm2026}.
DiffPT overfits (reconstruction w/ single lighting) but fails to generalize (relighting).}
\label{tab:multi_view_reconstruction_merged}
\centering
\setlength{\tabcolsep}{3pt}
\renewcommand{\arraystretch}{1.0}
\small
\begin{tabular}{@{}llccc|ccc|ccc@{}}
\toprule
& & \multicolumn{3}{c}{Real recon (136 img)} & \multicolumn{3}{c}{Synthetic recon (256 img)} & \multicolumn{3}{c}{Synthetic relight (2k img)} \\
\cmidrule(lr){3-5}\cmidrule(lr){6-8}\cmidrule(lr){9-11}
Geom. & Method
& PSNR$\uparrow$ & SSIM$\uparrow$ & LPIPS$\downarrow$
& PSNR$\uparrow$ & SSIM$\uparrow$ & LPIPS$\downarrow$
& PSNR$\uparrow$ & SSIM$\uparrow$ & LPIPS$\downarrow$ \\
\midrule
Known & DiffPT
& \sota{30.67} & \sota{0.959} & \sota{0.0307} %
& 27.50 & 0.929 & 0.0739 %
& 24.90 & 0.903 & 0.0911 \\

(ref.) & Ours
& 29.03 & 0.947 & 0.0331    %
& \sota{31.24} & \sota{0.960} & \sota{0.0370}    %
& \sota{30.46} & \sota{0.950} & \sota{0.0401} \\ %
\midrule
Recon. & LSRM
& 27.75 & 0.933 & 0.0388
& 22.82 & 0.872 & 0.1210 %
& 21.82 & 0.862 & 0.1241 \\

(LSRM) & Ours {\tiny (LSRM geo)}
& 27.01 & 0.936 & 0.0423
& 24.10 & 0.890 & 0.0896 %
& 23.71 & 0.885 & 0.0927 \\
\bottomrule
\end{tabular}
\end{table}

}

\newcommand{\tabSingleView}{

\begin{table}[tb]
  \caption{\textbf{Material generation.} Left: single-view to material generation. Right: text-guided material generation. All metrics evaluated on 32 scenes $\times$ 8 views $\times$ 8 probes.}
  \label{tab:matgen}
  \centering
  \setlength{\tabcolsep}{2pt}
  \renewcommand{\arraystretch}{0.9}
  \begin{small}
  \begin{tabular}{@{}lccc|lccc@{}}
    \toprule
    \multicolumn{4}{c}{Single-view to material} &
    \multicolumn{4}{c}{Text to material} \\
    \cmidrule(lr){1-4}\cmidrule(lr){5-8}
    Method & CLIP-FID~($\downarrow$) & CMMD~($\downarrow$) & LPIPS~($\downarrow$)
    & Method & CLIP-FID~($\downarrow$) & CMMD~($\downarrow$) & LPIPS~($\downarrow$) \\
    \midrule

    Hunyuan3D~2.1 %
    & 3.419 & 0.0437 & 0.0520
    &  & & & \\

    VideoMatGen %
    & 2.973 & 0.0236 & 0.0527
    & VideoMatGen
    & 4.725 & 0.0330 & \subsota{0.0639} \\

    Trellis.2 %
    & \subsota{2.227} & \subsota{0.0184} & \subsota{0.0395}
    & VideoMat
    & \subsota{4.376} & \subsota{0.0247} & 0.0652 \\
    
    Ours
    & \sota{1.520} & \sota{0.0081} & \sota{0.0325}
    & Ours
    & \sota{3.338} & \sota{0.0230} & \sota{0.0542} \\
    \bottomrule
  \end{tabular}
  \end{small}
\end{table}

}

\newcommand{\tabMatComponent}{

\begin{table}[tb]
  \caption{\textbf{Material components}. 
  We compute metrics on g-buffer renderings for 32 scenes $\times$ 17 views 
  from multi-view guidance in our BlenderVault test set. 
  We report scale-invariant PSNR for the base color renderings, and standard PSNR
  scores for base color, roughness, and metallicity g-buffer renderings.
  }
  \label{tab:mat_comp}
  \centering
  \setlength{\tabcolsep}{2pt} %
  \begin{small}
  \begin{tabular}{@{}lcccc@{}}
    \toprule
    Method & \multicolumn{2}{c}{Base color}  & Roughness & Metallicity \\
           & siPSNR~($\uparrow$) & PSNR~($\uparrow$) & PSNR~($\uparrow$) & PSNR~($\uparrow$) \\
    \midrule
    Ours           & \sota{29.19} & \sota{27.00} & \sota{24.32} & \sota{18.24}  \\
    DiffPT~\citep{hasselgren2022nvdiffrecmc}        & 25.08        & 20.21	    & 17.98 & 13.23 \\
    LSRM~\citep{lsrm2026}          & 23.17        & 15.46	    & 13.18 & 14.73 \\
    \bottomrule
  \end{tabular}
  \end{small}
\end{table}
}

\newcommand{\tabCRF}{

\begin{table}[htb]
    \caption{\textbf{Results Without CRF Adjustment}. 
        We provide quantitative results for \Cref{tab:multi_view_reconstruction_merged} without adjusting the CRF. 
        The tendency is the same, however some metrics become heavily biased due to the decomposition ambiguity. 
    }
    \label{tab:no_crf_multi_view_reconstruction_merged}
    \centering
    \setlength{\tabcolsep}{3pt}
    \renewcommand{\arraystretch}{1.0}
    \small
    \begin{tabular}{@{}llccc|ccc|ccc@{}}
    \toprule
    & & \multicolumn{3}{c}{Real recon (136 img)} & \multicolumn{3}{c}{Synthetic recon (256 img)} & \multicolumn{3}{c}{Synthetic relight (2k img)} \\
    \cmidrule(lr){3-5}\cmidrule(lr){6-8}\cmidrule(lr){9-11}
    Geom. & Method
    & PSNR$\uparrow$ & SSIM$\uparrow$ & LPIPS$\downarrow$
    & PSNR$\uparrow$ & SSIM$\uparrow$ & LPIPS$\downarrow$
    & PSNR$\uparrow$ & SSIM$\uparrow$ & LPIPS$\downarrow$ \\
    \midrule
    Known & DiffPT
    & \sota{27.01} & \sota{0.959} & \sota{0.0308} %
    & 21.99 & 0.908 & 0.0854 %
    & 21.14 & 0.903 & 0.0911 \\
    
    (ref.) & Ours
    & 22.66 & 0.930 & 0.0430    %
    & \sota{30.17} & \sota{0.958} & \sota{0.0327}    %
    & \sota{29.89} & \sota{0.954} & \sota{0.0350} \\ %
    \midrule
    Recon. & LSRM
    & 24.56 & 0.926 & 0.0445
    & 20.08 & 0.848 & 0.1296 %
    & 19.80 & 0.847 & 0.1330 \\
    
    (LSRM) & Ours {\tiny (LSRM geo)}
    & 22.22 & 0.923 & 0.0496
    & 23.61 & 0.884 & 0.0860 %
    & 23.43 & 0.889 & 0.0880 \\
    \bottomrule
    \end{tabular}
    \end{table}

}

\newcommand{\tabRuntime}{
\begin{table}[htb]
   \caption{\textbf{Runtime cost}. 
   Runtime cost and Peak VRAM usage.
   Scores are averages over four examples from the DTC dataset, measured on 
   an NVIDIA GB300 GPU.}
   \label{tab:runtime}
   \centering
   \begin{small}
   \begin{tabular}{@{}lcccccc@{}}
    \toprule
     Method  & Resolution & 2K recon & 8K super-res & Projection & Combined time & Peak VRAM  \\
     \midrule
     17-view  & 2K & 130~s & - & - & 2~min 10~s & 22 GiB \\
     117-view & 2K & 2610~s & - & - & 43~min 30~s & 48 GiB \\
     17-view  & 8K & 130~s & 267~s & 10~s & 6~min 47~s & 22 GiB \\
     117-view & 8K & 2610~s & 279~s & 10~s & 48~min 19~s & 48 GiB \\
     \bottomrule
   \end{tabular}
   \end{small}
 \end{table}
}

\newcommand{\tabRopeAblation}{
\begin{table}[tb]
  \caption{
    \textbf{RoPE embedding ablation.} We ablate our proposed 3D-aware rotary positional embedding against the standard Wan~2.1 RoPE on the text to material pipeline in Figure~\ref{tab:matgen} of the main paper, evaluated on the same 32 scenes $\times$ 8 views $\times$ 8 probes with material textures generated at a resolution of $512 \times 512$ texels. 
  }
  \label{tab:rope_ablation_sup}
  \centering
  \setlength{\tabcolsep}{6pt} %
  \renewcommand{\arraystretch}{0.9} %
  \begin{small}
  \begin{tabular}{@{}lccccc@{}}
    \toprule
    RoPE & Inputs & $G_{\mathrm{buf}}$ & CLIP-FID$\downarrow$ & CMMD$\downarrow$ & LPIPS$\downarrow$ \\
    \midrule
    Wan~2.1 & $\mathbf{p}_{uv}\!\!+\!\!f_{\mathrm{id}}$ & \redcross   & 4.036 & 0.0405 & 0.1005 \\
    Wan~2.1 & $\mathbf{p}_{uv}\!\!+\!\!f_{\mathrm{id}}$ & \greencheck & 3.677 & 0.0289 & 0.0971 \\
    3D-aware & $\mathbf{p}_{xyz}\!\!+\!\!f_\mathrm{id}$  & \greencheck & \sota{3.561} & \sota{0.0237} & \sota{0.0957} \\
    \bottomrule
  \end{tabular}
  \end{small}
\end{table}
}

\newcommand{\tabUpscaler}{

\begin{table}[tb]
  \caption{
    \textbf{Texture upscaler.} Results for $4\times$ texture upscaling comparing our method with popular alternatives. Results are quality metric averages for all textures in our synthetic dataset (32 objects). 
  }
  \label{tab:upscaler}
  \centering
  \setlength{\tabcolsep}{2pt} %
  \renewcommand{\arraystretch}{0.9} %
  \begin{small}
  \begin{tabular}{@{}lccccc@{}}
    \toprule
     & \multicolumn{2}{c}{Basecolor} & \multicolumn{2}{c}{(Rgh, Met)}\\
    Method & PSNR~($\uparrow$) & LPIPS~($\downarrow$) & PSNR~($\uparrow$) & LPIPS~($\downarrow$)\\
    \midrule
    Our        & \sota{34.7} & \sota{0.061} & \subsota{38.0} & \sota{0.025} \\
    PBR-SR~\citep{chen2025pbrsr} & 30.9 & 0.151 & \sota{40.6} & \subsota{0.030} \\
    ESRGAN~\citep{wang2018esrgan} & \subsota{32.6} & \subsota{0.146} & 37.9 & 0.041 \\
    \bottomrule
  \end{tabular}
  \end{small}
\end{table}

}

{%
\maketitle
    \centering
    \includegraphics[width=\textwidth]{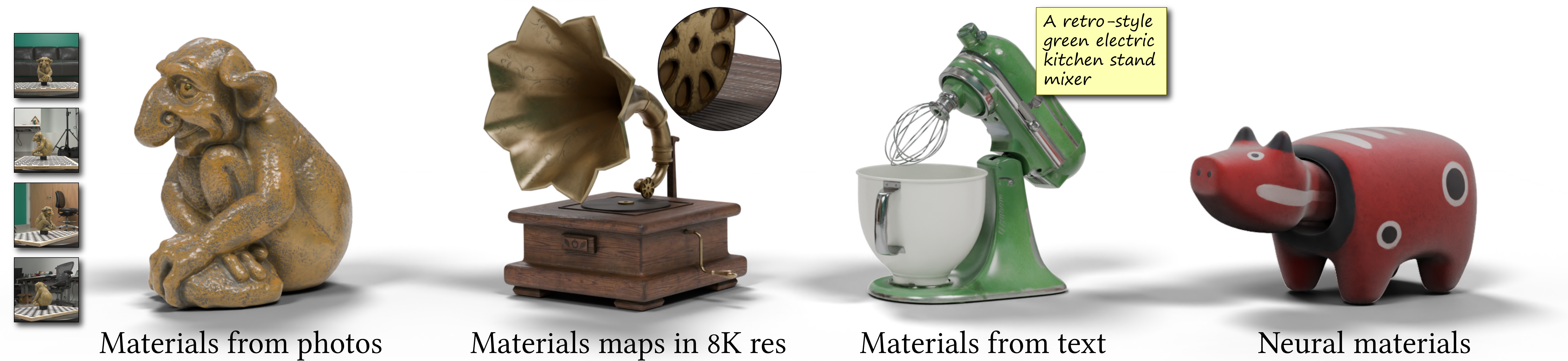}
    \captionof{figure}{Our texture space diffusion model generates PBR materials from image- or text conditioning, scales to 8K material maps and generalizes to neural material representations\vspace{1em}.}
    \label{fig:teaser}
}

\begin{abstract}
We present a method for generating high quality materials for 3D objects entirely in texture space. We finetune a video diffusion transformer for text-guided material generation, multi-view material generation, and material upscaling. Our key insight is to use the known projection from image space to texture space, enabling the diffusion process to generalize across arbitrary geometries and texture parameterizations. This approach also avoids the view consistency issues inherent in video and multi-view diffusion models. Because texture space is two dimensional, we can reuse the strong priors of pretrained video diffusion models. We apply our method to high quality material reconstruction from posed photos captured under unknown lighting, as well as to text- and image guided material generation. Our method can scale to high resolutions (8K), 100+ input views, and neural material representations. In quantitative and qualitative evaluations we show state-of-the-art results for material generation and reconstruction.
\end{abstract}

\addtocontents{toc}{\protect\setcounter{tocdepth}{-1}}
\section{Introduction}
\label{sec:intro}

In this work, we focus on the task of generating realistic materials for 3D objects from text descriptions and/or photographs.
A common way of representing realistic materials for a 3D object is storing the spatially varying material
parameters in texture maps.
This is the standard approach for materials in 3D graphics systems, e.g., Blender and Unreal Engine. However, authoring detailed materials on complex 3D assets is a labor-intensive task. 

Recent advances in generative modeling have made it increasingly practical to synthesize 3D geometry and appearance from text or images. Existing approaches broadly fall into two categories: \emph{image-space} and \emph{object-space} models. 

Image-space methods~\citep{poole2022dreamfusion,deng2024flashtex,Zhang2024dreammat,youwang2024paintit,munkberg2025videomat,hasselgren2026videomatgen} build on image or video generative models to produce multiple views of an object, 
from which geometry and materials are reconstructed using optimization through differentiable rendering.
They are mainly limited by view consistency: the same surface point may be predicted differently across generated 
views, often leading to reduced details, suppressed specular highlights or blurred textures.

Object-space methods~\citep{xiang2025trellis2,he2025materialmvp,wu2025direct3ds2gigascale3dgeneration, yu2024texgen,li2026pixal3d, lsrm2026} instead rely on three-dimensional representations such as sparse voxel grids or triplanes.
Because features are attached to fixed locations, these methods avoid view-consistency issues.
However, sparse 3D representations are memory- and compute-intensive, require specialized complex architectures, and are 
trained from scratch on synthetic data rather than leveraging the priors of large video foundation models, which may lead to lower visual quality.

In this work, 
inspired by TEXGen~\citep{yu2024texgen}, 
we explore the \emph{texture space} domain for high quality material generation. 
Texture space is two-dimensional, allowing us to easily adapt existing image and video models, and is view consistent by design.
 In Figure~\ref{fig:texspace}, we illustrate the concept of texture mapping, and the projection from views back into texture space.
Our main target application is material estimation for a known, scanned, object 
with uncontrolled lighting. 
Each view is projected into a shared UV atlas, and a generative video model is conditioned 
on the shaded texture space views, generating output frames with the physically-based rendering (PBR) material parameters: (\emph{basecolor}, \emph{height}, \emph{roughness}, \emph{metalness}).

We show that our method can be made robust to inconsistencies in both input views and geometry, making it a versatile component 
in larger content-generation pipelines. The input views can include high quality photos from a controlled setup, phone captures, or even views generated by a (e.g. depth-conditioned) video model. The geometry can be an accurate scan, or a coarser extraction from recent image-conditioned models~\citep{lsrm2026,xiang2025trellis2,sam3d2025}.

\figTexSpace

We complement our \emph{multi-view to material} model with a generative upscaler trained on UV-atlases of material parameters, which, combined with inference time augmentation, 
can be used to scale to both higher resolution (8K) and more input views (100+) than used in training.
Finally, we show that text-conditioned and single-image conditioned material generations are feasible in texture space by finetuning a diffusion transformer model with a novel 3D-aware rotary positional embedding which encodes the (per-texel) world space position and frame ID. Our main contributions are:

\begin{itemize}
    \item To our knowledge, the first diffusion framework that generates complete PBR materials entirely in texture space, combining a view-consistent 2D representation with pretrained image and video priors.
    \item We show that full texture-space generation enables robust material reconstruction under unknown lighting from single or multiple views.
    \item Training-free noise rolling and coverage-aware expert aggregation scale inference to 8K textures and over 100 input views.
\end{itemize}

\section{Related Work}

\figSystem

\paragraph{Image and Video Diffusion.}
Image diffusion models generate samples through iterative denoising~\citep{sohl2015deep, ho2020denoising, dhariwal2021diffusion}.
Video diffusion models~\citep{blattmann2023videoldm, blattmann2023svd, hong2023cogvideo, yang2024cogvideox, cosmos_short,wan2025wan} extend this concept to the temporal domain. 
Diffusion transformers (DiTs)~\citep{peebles2022dit} have become a common
architecture due to their performance and flexible finetuning opportunities.
In this work, we build upon the DiT-based Wan~2.1 model~\citep{wan2025wan} and use its temporal
representation to process multi-view observations and produce multiple modalities.

\paragraph{Diffusion-based 3D Asset Generation.}
Many methods build on image diffusion models with \emph{score distillation sampling} (SDS) to produce complete 3D assets~\citep{poole2022dreamfusion,zhu2023hifa,wang2023prolificdreamer,zhu2023hifa}. 
SDS-based methods require slow optimization, prompting the development of methods that reconstruct in a single forward pass using a pretrained transformer~\citep{li2023instant3d,zhang2024gslrm}.

A common limitation of image models is the lack of view consistency, which may show up as blur in the extracted textures. 
Multi-view diffusion models e.g., MVDream
\citep{shi2023MVDream} address this by jointly generating images from several
canonical viewpoints.
SV3D~\citep{voleti2024sv3d} and Hi3D~\citep{yang2024hi3d} further improve on this aspect by finetuning \emph{video diffusion models} for object rotations, and extract 3D models from the generated views. However, these approaches have limited resolution and do not provide Physically Based Rendering (PBR) materials. 

Another direction is to apply the generative process in 3D space, which is view-consistent by design.
3DTopia-XL~\citep{chen2024primx} and Trellis~\citep{xiang2024structured,xiang2025trellis2} encode the 3D shape, textures, and materials in volumetric primitives anchored and jointly generates shape and PBR materials. Recently, LSRM~\citep{lsrm2026} leverages sparse attention to extend this approach to higher resolutions.
These methods target complete asset generation, whereas we focus on
PBR materials.

\paragraph{Material Reconstruction.}
Material reconstruction is commonly formulated as an inverse-rendering
problem from dense set of inputs. 
Differentiable rasterization~\citep{Laine2020diffrast} has been successfully applied to photogrammetry~\citep{Munkberg_2022_CVPR}. 
Differentiable path tracing~\citep{Zhang:2020:PSDR,Mitsuba3} accurately simulates global 
illumination effects, and has higher potential reconstruction quality~\citep{hasselgren2022nvdiffrecmc}, but introduces Monte-Carlo noise in the training process, which makes gradient-based optimization more challenging. 
Volumetric scene representations such as NeRFs~\citep{Mildenhall2020} and 
Gaussian splatting~\citep{kerbl3Dgaussians} provide a smoother optimization landscape with impressive novel-view synthesis quality, but disentangling shape, materials, and lighting is non-trivial. 

A closely related line of research is intrinsic decomposition of images, including per-pixel material parameter estimation from sparse set of inputs. 
To account for the decomposition ambiguity, recent methods use image or video diffusion models to generate materials ~\citep{kocsis2023iid,chen2024intrinsicanything,zeng2024rgb,zhu2024mcmat,DiffusionRenderer,ying2025chord,zeng2026rgbxnext,luo2026matpedia} and potentially combine with optimization-based inverse rendering~\citep{kocsis2026iif,munkberg2025videomat}.

Recent works leverage diffusion models for generating more expressive \emph{neural} materials on 2D patches~\citep{raghavanmullia2025genneumat,xue2026videoneumat}, on 3D geometry~\citep{yu2026enhance} or to extract neural materials from images~\citep{youwang2026neumatex}. 
These methods are operating in image-space, while we propose to directly generate in texture space. 
While we focus on PBR materials in this paper, neural materials are an exciting future direction, and we include a proof-of-concept example.

\paragraph{Diffusion-based Material Generation.}
Earlier approaches formulate the task of material generation as retrieval or interpolation of a database of known 
high-quality materials or material graphs, and learn to project the input (image or text) onto the known representation~\citep{zhang2024mapa, ceylan2024matatlas,fang2024makeitreal}. 
These methods are limited by the expressiveness of their material databases, but are well-regularized.

Hybrid approaches combine image diffusion models with material distillation using inpainting, or coarse-to-fine texture  refinement~\citep{richardson2023texture,chen2023text2tex,zeng2024paint3d,youwang2024paintit,hadadan2025}. 
Recent methods fine-tune the diffusion model to condition on geometry or lighting ~\citep{Zhang2024dreammat,deng2024flashtex} or even to directly generate the intrinsic properties from text or known geometry ~\citep{kocsis2025intrinsix,munkberg2025videomat,hasselgren2026videomatgen}.
CLAY's~\citep{zhang2024clay} material generation model uses a finetuned multi-view image diffusion model~\citep{shi2023MVDream} conditioned on normal maps. The material model generates four canonical views of the PBR texture maps, which are then projected into texture space. Several recent methods~\citep{,huang2024materialanything,feng2025romantex,boss2025sf3d,he2025materialmvp,shao2025mvpainter,yang2025pandora3d,engelhardt2025svim3d,seed3d} 
extend this approach with additional input conditioning (normal, depth and/or world space positions).
However, these methods operate in image space, potentially leading to texture seams and limited resolution. 

TEXGen~\citep{yu2024texgen} shows that it is possible to directly generate in texture space. 
To provide geometry information, they interleave convolutions in texture space with attention layers on point clouds to outpaint diffuse textures. 
Inspired by this, we formulate the generation process entirely in texture space that can generate complete PBR materials.

\section{Method}

Our texture space diffusion method, as shown in \cref{fig:system}, predicts physically-based rendering (PBR) material textures~\citep{burley2012physically,karis2013real} from 
posed views. We assume a given 3D model with a valid texture parameterization (but no textures) as input, alongside a text prompt describing the material. Our approach is multi-modal, and can predict 
materials from: 1) one or multiple posed views, 2) a text prompt, or 3) low-resolution material maps. 

Given known camera parameters and geometry, we project the views into texture space and formulate
the diffusion process in texture space, leveraging pre-trained 2D diffusion priors (e.g., a video diffusion model).
One can see our approach as a texture completion task, conditioned on world space positions, normals, and a text prompt. Our finetuned model outputs two images, representing the 
PBR material maps in texture space, one for base color and the other encoding a height map, roughness, and metallicity combined into an RGB image. 
We use the height map only for surface normal perturbation (bump mapping), but it could also be used for displacement mapping.

\subsection{Architecture}

\label{sec:video_model1}

Our input tensor, $\videoInput$, consists of texture-space views for N frames, alongside the world space positions and normals, also in texture space. 
We finetune a recent Diffusion Transformer (DiT) video model, Wan2.1-1.3B~\citep{wan2025wan}, to generate complete PBR material maps. 

The model comprises a VAE encoder-decoder pair, $(\vaeEncoder, \vaeDecoder)$, and a transformer-based denoising function, $\diffusionModelFn$.
First, the input tensor, $\videoInput$, is encoded into a latent tensor, $\textbf{z}^{\videoInput} = \vaeEncoder(\videoInput)$.
Similarly, the shared conditions, $\textbf{C}$, (texture space world space positions, and world space normals) are encoded into a latent tensor $\textbf{z}^{\textbf{c}}$. The target latent variable, $\textbf{z}_0^{\matmap}$, is constructed by encoding the reference base color, and HRM (height, roughness, metallicity) using the $\vaeEncoder$ encoder.
Noise, $\diffusionNoise$, is introduced to our latent, $\textbf{z}_0^{\matmap}$, representing the material parameters, to produce $\textbf{z}_\tau^{\matmap}$. 

We use the flow matching training objective, in which the forward process is defined 
as a linear interpolation between the clean data and noise: $\textbf{z}_\tau^{\matmap} = (1-\tau) \textbf{z}_0^{\matmap} + \tau \epsilon$. 
In flow matching, $\diffusionModelFn$ is trained to predict the target velocity 
field, where "velocity" is formulated 
as $v = \epsilon-\textbf{z}_0^{\matmap}$.

The model parameters, $\theta$, of the diffusion model, $\diffusionModelFn$, are optimized by minimizing the objective function: 
\begin{equation}
\mathcal{L}(\theta) = \mathbb{E}_{\textbf{z}_0^{\matmap}\sim\dataDistribution,\diffusionNoise\sim \mathcal{N} (0,\sigma^2 I)} \left\| \diffusionModelFn ([\textbf{z}_\tau^{\matmap},\textbf{z}^{\videoInput}, \textbf{z}^{\textbf{c}}]; \typeEmb, \tau) - (\epsilon-\textbf{z}_0^{\matmap}) \right\|_2^2
\label{eq:objective},
\end{equation}
where $[\cdot]$ denotes concatenation along the temporal dimension and $\typeEmb$ is the encoded text prompt (encoded using T5-XXL~\citep{raffel2023t5}). We learn unique \emph{token type encodings}~\citep{zeng2026rgbxnext} for each of $\textbf{z}_\tau^{\matmap}$,$\textbf{z}^{\videoInput}$, $\textbf{z}^{\textbf{c}}$. The target latent, input frames, and conditions comprise a sequence of latent frames, and the DiT can learn to attend to each frame in this sequence.

We finetune all DiT layers for 15k iterations on 32 A100 GPUs, using gradient accumulation over 4 steps, which results in an effective batch size of 128, and progressively increase the training resolution from $512^2 \rightarrow 1024^2 \rightarrow 2048^2$.

\paragraph{Multi-view to material generation}

The task is to merge a set of sparse texture space (shaded) views into dense, 
fully covered texture space material maps. The model needs to learn how to demodulate lighting from the
texture space views, merge the observations, and hallucinate plausible detail in regions with missing coverage.
Here, $\videoInput$ consists of $N=17$ texture space views,
with spatial resolution $H \times W$. These are encoded using  $\vaeEncoder$ into a latent tensor, $\textbf{z}^{\videoInput}$. 

Our training data consists of path traced views of 3D models.
However, real captures and current video diffusion models come with view-inconsistencies.
To make our model more robust, we augment the training data with random corruptions of the input views. 
First, we create a version of the dataset where we add Gaussian noise to the views and denoise them with
FLUX.1-dev~\citep{flux2024} img2img (using strength $\in [0, 0.3]$, guidance scale $\in [0, 3.0]$ and 20 denoising steps). 
Secondly, we add Gaussian blur with randomized $\sigma \in [0, 15]$ to the input views during training. 
These augmentations encourage our material extraction model to be robust to imperfections, while still producing high frequency details.

\paragraph{Single-view to material generation}

In this setting, we only have a single observation of the materials in a shaded view, which we project into texture space. The model, guided by the texture space sparse input and conditions, needs to demodulate lighting and inpaint plausible details in missing regions. Here $\videoInput$ consists of a single, sparse texture space view.
To improve consistency of the generated material, inspired by RomanTex~\citep{feng2025romantex}, we introduce a novel \emph{3D-aware} rotary positional embedding (RoPE), combining the frame ID and world space position. 
To our knowledge, we are the first to apply a 3D-aware RoPE in texture space.
Encoding world space positions directly in RoPE supplies a stronger or more direct geometric bias than G-buffer conditioning alone, improving attention in spatially adjacent surface regions that are distant in texture space.
In our supplemental material, we ablate the impact of this 3D-aware RoPE.

 \paragraph{Text to material generation}

The text to material model is near identical to the single-view model.
We replace $\videoInput$ with a binary UV coverage mask, indicating which parts of texture space are populated or empty, 
and use the same 3D-aware RoPE.

\paragraph{Texture space upscaling and refinement}

To increase the resolution of the generated material maps, we propose a texture-space generative refinement model. This model is primarily useful to complement the multi-view to material generation method of 
\cref{sec:video_model1}. 
The upscaler model is similar to the single-view to material model, but with the single frame replaced with two low resolution PBR material maps (with full coverage). In training, the low resolution PBR material guides are generated by first downscaling then upscaling the reference materials with area weighted or nearest neighbor filtering (randomly selected) using the scaling factors: 2,4,8, and 16.

\paragraph{Neural materials} 

The above methods can be extended to handle neural material representations by replacing the target latent variable, $\textbf{z}_0^{\matmap}$ (which consists of encoded PBR material maps), with VAE-encoded neural material textures.
As a proof of concept, we apply the dataset and neural material representation of \citet{yu2026enhance} to our generative 
image-conditioned model. Neural representations can be used to encode more complex appearances, such as the fuzzy material shown in our result in the right column of \cref{fig:teaser}. 

\subsection{Training Dataset}
\label{sec:dataset}
Our dataset consists of 120k videos of object-centric renderings of 3D models from TexVerse~\citep{zhang2025texverse}.
For each object, we render a video with 17 frames at a resolution of 2048$\times$2048, using a path tracer with three bounces, black background and Blender AgX tonemapping. 
For lighting, we use 697 light probes from Poly Haven~\citep{polyhaven}, one randomly selected for each object.
Each frame is projected into texture space following~\cref{fig:texspace}. 
We automatically generate captions using Qwen2.5-VL-7B~\citep{qwen25}. 
We also rasterize intrinsic conditioning guides (world space positions and world space normals) directly in texture space and store them alongside reference base color, roughness, metallicity and height textures.
The height map is not available for most assets, and we reconstruct it from the normal map using standard conversion tools.

\subsection{Inference Scaling}
\label{sec:inference_scaling}

Scaling transformer models to higher resolution or more input views is challenging due to attention cost. We show that inference-time methods can enable scaling beyond the training domain. 

\paragraph{Increasing texture resolution}

Our high-resolution inference uses progressive noise rolling, inspired by \citet{bartal2023multidiff,vecchio2023controlmat}. 
We first run inference on the original 2K resolution. We upscale the resulting image to 8K, add a small amount of noise, and resume denoising from that point. 
In each diffusion step, we split all model inputs ($\textbf{z}_\tau^{\matmap},\textbf{z}^{\videoInput}, \textbf{z}^{\textbf{c}}$) into non-overlapping 2K crops, evaluate $\diffusionModelFn$ on each crop, and stitch the result 
together. We hide seams by applying a random spatial roll in each diffusion step, thus randomly offsetting the crops.
We apply noise rolling in the upscaling step for best runtime performance.

\paragraph{More input views}
To use more input observations, we evaluate them batch-wise and propose a fusing strategy. 
We split the input views into overlapping batches.
In each denoising step, we predict the flow for all batches, given the same noisy latents, referred to as expert predictions. 
To aggregate the predictions, we compute per-texel coverage scores for all experts. 
These coverage maps can act as an uncertainty-measure, i.e. more views observe a texel, the more certain the prediction should be. 
For each texel, we keep the top-2 expert predictions and use their coverage-weighted average. 
This retains redundancy, while avoiding the oversmoothing that would result from averaging all experts.

\figDTCmain

\section{Experiments}
\label{sec:experiments}

We evaluate our method against the state-of-the-art in multi-view to material reconstruction and text/image conditioned material generation. Additionally, we ablate robustness to view- and geometry inconsistencies, and our inference-time methods for increasing the number of input views and resolution. For quantitative results we use a synthetic dataset of 32 held out 3D models from the BlenderVault dataset~\citep{litman2025materialfusion}. 

\subsection{Multi-view to materials} 
\label{sec:res_multi_view}

We evaluate our multi-view material reconstruction model on two test sets: 1) A fully controlled synthetic test set with 32 examples 2) Eight real multi-view captures from the DTC~\citep{dong2025dtc} dataset. 
For the synthetic set, we render each 3D model with extracted materials from eight viewpoints with the original lighting and in eight \emph{novel} lighting conditions. For the DTC examples, we use the original lighting in rendering. Error metrics are computed as averages over all images.

To account for any tonemapping bias of the baselines, we fit a parametric camera response function (CRF), following ~\citet{lin2025iris}, for each series based on ground truth renderings under the original lighting condition. 
Material estimation under unknown lighting is fundamentally ambiguous as observed intensity depends jointly on illumination intensity and material reflectance. 
Consequently, the different methods produce materials with different systematic bias in the material parameters, which can be largely reduced with CRF adjustment.
For completeness, we include series without CRF adjustments in our supplementary material. 

\tabMultiview

In \cref{tab:multi_view_reconstruction_merged} we report metrics for multi-view reconstruction for synthetic and real examples. As baselines, we include DiffPT~\citep{hasselgren2022nvdiffrecmc}, an optimization-based approach through differentiable path tracing, and LSRM~\citep{lsrm2026}, a large reconstruction model for geometry and material estimation. In the left part, we report metrics for 
renderings of the reconstructed materials with the \emph{original lighting}.
We note that our method (with reference geometry) has the highest scores on all metrics for the synthetic set. For the real (DTC) examples, the optimization-based approach, DiffPT, which overfits to this configuration, has slightly better metrics. 
In the right column, we evaluate the reconstructed materials across eight novel lighting configurations. The scores indicate that DiffPT fails to properly disentangle lighting and material information, as also shown in \cref{fig:dtc}. LSRM has slightly lower scores, but solves the more challenging task of jointly reconstructing the geometry and materials. We include a series with our method using LSRM geometry, which improves slightly over LSRM across all metrics in the multi-illumination setting. 

To ablate robustness to imperfect views, we leverage a pre-trained video model (Wan2.2-VACE-Fun-A14B)~\citep{vace2025} to generate a 360 degree orbit of the 3D object, conditioned on a single image and a depth video. Here, we use a smaller test set of 17 objects, 
as the depth-conditioned video model struggles to generate plausible orbits for rotationally symmetrical objects.
In \cref{fig:vace}, we observe that our method robustly handles view imperfections, and can produce plausible materials 
from views synthesized by a video model.

In \cref{fig:teapot}, we show one example of the impact of our inference augmentations, which allows us
to scale the material maps to 8K resolution  while leveraging more input views. More results are shown in the supplement. 

\figVACEsmall

\subsection{Generative materials} 

Generative material models hallucinate material parameters from text or from a shaded example image. In \cref{tab:matgen} (left) we evaluate our image conditioned method against Trellis.2~\citep{xiang2025trellis2}, Hunyuan3D 2.1~\citep{hunyuan3d2025}, and VideoMatGen~\citep{hasselgren2026videomatgen}. 
In all cases, we run the pipelines with known geometry, only extracting the materials. This mode is directly 
supported by their respective code releases.
We evaluate text to material generation in \cref{tab:matgen} (right). Here we compare against VideoMat~\citep{munkberg2025videomat} and VideoMatGen which have native
support for text to material.

\cref{fig:single_view} shows visual examples of single-view to material generations from our synthetic test set. Our method is robust over a large variety of objects and lighting conditions, producing semantically meaningful results over all test objects. In \cref{fig:texgen_main} we compare against TEXGen~\citep{yu2024texgen} on the single-view to \emph{diffuse} texture generation task. Please refer to the supplemental material for comprehensive visual examples and quantitative results of text-guided material generation.

\figSingleView
\tabSingleView

\figTEXGENmain

\figInfAblation

\section{Limitations and Future Work}

Our implementation is currently not optimized. Inference is costly, in particular for the multi-view model: 130~s for the 17-view model @2K on a GB300 GPU (see supplement). Recent video model acceleration and distillation techniques offer promising directions for improving efficiency.

Capturing specular appearance is challenging, and while our approach 
improves over previous work in this aspect, careful data curation can likely help. We are also excited about looking closer at neural material representations, which accurately model specular appearance~\citep{yu2026enhance}.

Wan-2.1~\citep{wan2025wan} operates on patches of $16\times 16$, which is the smallest granularity for attention and RoPE. Consequently, in patches with multiple overlapping texture atlas segments our model lacks the information to distinguish them. Although higher texture resolutions mitigate this issue, patch-aware texture unwrapping, or pixel-space diffusion models are promising directions for future work.

\section{Conclusion}

We show that texture space diffusion models can be powerful tools for material reconstruction and generation, achieving on par or higher quality compared to the state of the art. Operating in texture space ignores the common problems of view consistency or data sparsity, making our method both simple and scalable.

\subsubsection*{Acknowledgments}
We thank Kim Youwang, Zhengqin Li, Nicholas Sharp, Steve Marschner, Zheng Zeng, Andrea Weidlich, Blaire Yu, and Milo\v{s} Ha\v{s}an for helpful discussions, feedback, and sharing code. 
We also thank Aaron Lefohn and Chris Wyman for supporting this research.

\newpage

\bibliography{main}
\bibliographystyle{iclr2027_conference}

\addtocontents{toc}{\protect\setcounter{tocdepth}{2}}
\newpage
\title{Texture Space Material Diffusion \\\vspace{2.0mm} --- Supplementary material ---}
\maketitle

\appendix

\newtheorem{assume}{Assumption}
\newtheorem{definition}{Definition}
\newtheorem{lemma}{Lemma}

\setcounter{section}{0}
\setcounter{figure}{0}
\setcounter{table}{0}
\setcounter{equation}{0}

\renewcommand\thesection{\Alph{section}}
\renewcommand\thefigure{S\arabic{figure}}
\renewcommand{\thetable}{S\arabic{table}}
\renewcommand\theequation{\alph{equation}}

\makeatletter
\renewcommand{\numberline}[1]{#1\hspace{1.0em}}
\makeatother

\begingroup
\hypersetup{linkcolor=black}
\makeatletter
\renewcommand{\l@section}[2]{\addpenalty{-\@highpenalty}%
  \addvspace{2pt}\@dottedtocline{1}{0em}{1.5em}{\bfseries #1}{\bfseries #2}}
\renewcommand{\l@subsection}{\@dottedtocline{2}{1.2em}{2.3em}}
\makeatother
\noindent\rule{\linewidth}{1.0pt}
\vspace{-6mm}
\tableofcontents
\noindent\rule{\linewidth}{1.0pt}
\endgroup

\section{Implementation details}
\label{sec:sup_implementation}

\subsection{CRF adjustment}
\label{sec:sup_crf}
Appearance decomposition is a highly ambiguous task, making quantitative evaluations challenging. 
To account for the ambiguity, we fit a parametric camera response function (CRF) and exposure value for all the multi-view methods presented in the paper.
We parametrize the response function using \citet{grossberg2004emor} and fit it with a rerendering loss ~\citep{liu2020crf}.
Specifically, for all the available views, we render the object in HDR using the original lighting condition and the predicted materials. 
Then, we start an alternating optimization between exposure and CRF fitting, giving a total of 10 trainable parameters (1 exposure and 3 curve parameters for each color channel). 
We apply the exposure, then the learned CRF to tonemap the image and compare against the original image using L1 loss. 
We tune the exposure for 25 steps, then the CRF parameters for 200 steps and alternate for 20 cycles using learning rate of 0.01. 
Following IRIS ~\citep{lin2025iris}, we apply a monotonicity regularizer with weight of 0.1. 

\figCRF
\tabCRF

After the fitting, we save the converged parameters and apply them to tonemap all the renderings, including relightings. 
We visualize the effect of CRF adjustment in \Cref{fig:crf} and provide full quantitative results without CRF adjustment in \Cref{tab:no_crf_multi_view_reconstruction_merged}.

\subsection{Inference scaling}
\label{sec:sup_inf_scaling}
As described in the main text, we are combining our proposed super-resolution model with noise rolling. 
The super-resolution model is trained on $512 \times 512$ input and $2048 \times 2048$ output resolution, giving 4$\times$ upscaling. 
To scale this further to handle $2048 \times 2048$ inputs, we apply progressive noise rolling. 
First, we generate the texture in native $2048 \times 2048$ resolution, then upsample it with nearest neighbor interpolation.
We add noise according to timestep 0.2 and denoise with 50 steps using noise rolling. 
In \cref{tab:runtime} we report the runtime cost and peak VRAM usage for these generations.

\tabRuntime

\subsection{Diffusion transformer model} 
\label{sec:sup_transformer_model}
Our input tensor, $\videoInput$, consists of texture-space views for $N$ frames, alongside the world space positions and normals, also in texture space. World space positions and normals are quantized to 8 bits per channel and normalized to the [-1,1] range before the VAE encoding, $\vaeEncoder$.
The target latent variable, $\textbf{z}_0^{\matmap}$, is constructed by encoding the reference base color, and HRM (height, roughness, metallicity) as two RGB images using the $\vaeEncoder$ encoder, then concatenating the encoded images along the temporal dimension.

\subsection{3D-aware RoPE}
\label{sec:sup_impl_rope}
A rotary positional embedding (RoPE) which incorporates 3D positions was introduced in RomanTex~\citep{feng2025romantex}, which replaces a 2D RoPE encoding of pixel $xy$-coordinates
by a 3D-aware RoPE which encodes the 3D world space position. Positional Encoding Field~\citep{bai2025positional} proposes another 3D RoPE variant, combining pixel $xy$-coordinates and depth.
The motivation is to improve view consistencies in multi-view generation.

In this paper, we adapt a variant of this for our texture space diffusion framework, and modify the RoPE (combining pixel $xy$-coordinates, $\mathbf{p}_{uv}$ and frame id $f_{\mathrm{id}}$) of Wan~2.1
to a 3D-aware RoPE encoding (combining a 3D world space position and frame id: $\mathbf{p}_{xyz} + f_\mathrm{id}$) to inject stronger geometry guidance in our single-view and text-to-material generation pipelines.

The 3D position input for our RoPE is obtained by using the texture space view of the world space positions
(normalized to [0,1]) downsampled 16$\times$ spatially to match the spatial token size (8$\times$ reduction from VAE encoding and 2$\times$ more from the DiT's patchification). We downsample via masked pooling:

\begin{lstlisting}[language=Python] 
import torch.nn.functional as F
    
def downscale_masked_pool(col, mask, eps=1e-8): # weighted sum per tile
    col_sum = F.avg_pool2d(col * mask, kernel_size=16, stride=16)
    mask_sum = F.avg_pool2d(mask, kernel_size=16, stride=16) 
    return col_sum /  mask_sum.clamp_min(eps)
\end{lstlisting}

In the Wan2.1~1.3B version we are using, the attention head dimension is 128.
RoPE allocates frequencies to each input, and the standard split is 
(44,42,42) for ($f_\mathrm{id}$, $p_u$, $p_v$).
In our 3D-aware variant, we allocate the frequencies as (44,28,28,28) for ($f_\mathrm{id}$, $p_x$, $p_y$, $p_z$).
Each normalized world-position component is clamped to [0,1] and scaled over the full 1024-entry frequency table. We use the frequency table from Wan unmodified. 
We interpolate between neighboring complex frequency entries and renormalize the result to unit magnitude.
The same spatial world-position map is expanded over all frames and applied to both output and conditioning tokens. Frame identity still comes from the frame component.

During training, the standard and 3D-aware RoPE variants are linearly blended over the first 2,000 steps.

\subsection{Material height map}
\label{sec:sup_impl_height}
We predict material height maps to support surface details through surface normal perturbations, or bump mapping. 
The height map representation is convenient as it can be packed as an $(r,g,b)$-triplet with roughness and metallicity 
at no extra runtime cost, but extracted normals will depend on a user-adjustable height scale parameter. We follow the workflow of the AwesomeBump~\citep{awesomebump} normal map library and use a height scale of $\alpha=1$ throughout the evaluation. Assuming pixel coordinates with $y$ increasing upward, and a height field texture, $h(x,y)$, we compute the normal as:
\begin{equation}
\hat{\mathbf{n}}(x,y) = \operatorname{normalize}\left(
\alpha \left[h(x,y)-h(x+1,y)\right],\;
\alpha \left[h(x,y)-h(x,y+1)\right],\;
1
\right)
\end{equation}

We similarly extract height maps from normal maps when generating our dataset, as most objects specify normal maps rather than height. This is considerably more complex and requires solving a Poisson equation~\citep{queau2017nis}:
\begin{equation}
\nabla^2 h
=
-\frac{1}{\alpha}\left[
\frac{\partial}{\partial x}\left(\frac{n_x}{n_z}\right)
+
\frac{\partial}{\partial y}\left(\frac{n_y}{n_z}\right)
\right].
\end{equation}
Since the normal maps are not necessarily integrable, an exact height reconstruction may not exist. We use the approximate iterative solver of AwesomeBump, which is fast and did not introduce objectionable artifacts in the height map. Defining $s_x = n_x/(\alpha n_z)$ and $s_y = n_y/(\alpha n_z)$, we iteratively update the height field (50 iterations) using:
\begin{equation}
\begin{aligned}
h^{(k+1)}(x,y)
={}& \frac{1}{4}\Big[
h^{(k)}(x+1,y) + h^{(k)}(x-1,y) + h^{(k)}(x,y+1) + h^{(k)}(x,y-1)
\Big] \\
&+ \frac{1}{8}\Big[
s_x(x+1,y) - s_x(x-1,y) + s_y(x,y+1) - s_y(x,y-1)
\Big].
\end{aligned}
\end{equation}

\section{Evaluation protocol}
\label{sup:eval_protocol}

We evaluate our method using standard image-quality and perceptual metrics. For reconstruction tasks, including material extraction from multi-view captures, we report PSNR, SSIM, and LPIPS. For generative tasks, such as text-to-material and single-view material generation, we report distribution-based metrics CLIP-FID and CMMD, following common practice.

All metrics are computed on rendered images with a black background. For reconstruction tasks, we use a rendering resolution of $1024 \times 1024$ pixels to adequately capture high-resolution textures. We use a rendering resolution of $512 \times 512$ pixels for the distribution-based metrics, as they rely on CLIP which is limited in resolution (336$\times$336 pixels for CMMD, utilizing the \texttt{ViT-L/14@336px} vision model and 224$\times$224 pixels for CLIP-FID). We apply no image processing other than tonemapping, using either the ACES tonemapper or CRF adjustments as indicated.

Notably, the black background makes scores dependent on scene scale. For synthetic benchmarks, we use a camera path where the object bounding sphere is maximized but guaranteed to fit on screen. For the DTC~\citep{dong2025dtc} dataset, we pick representative views of a revolution where the object is large on screen.

Our protocol differs from the Stanford-ORB methodology~\citep{kuang2023stanfordorb} adopted by DTC~\citep{dong2025dtc} and LSRM~\citep{lsrm2026}. The Stanford-ORB protocol uses a lower resolution of $512 \times 512$ pixels and erodes the reference image mask by $5\times5$ pixels, inserting black background in both reference and rendered image in the eroded regions. Furthermore, they use two variants of PSNR, one (PSNR-L) applied to srgb-remapped clipped (to $[0,1]$) linear color values, and another (PSNR-H) applied to linear values which are clipped to $[0,4]$. In both cases, the images are normalized using a mean color normalization approach~\citep{physg2021}. The set of evaluation views are also different. As a consistency check, we evaluated the LSRM materials using the Stanford-ORB metrics pipeline and obtained scores comparable to those reported in the original paper.

\paragraph{Material Generation.}
For the single-view to material and text to material experiments, we run all algorithms with known reference geometry. Like our method, VideoMat and VideoMatGen require known geometry, and the released code of Trellis.2 (their ``PBR Texture Generation mode'') and Hunyuan3D 2.1 (the Hunyuan3DPaintPipeline), both support PBR material generation with known geometry. We use the same conditioning image for all algorithms.

\figComponents
\tabMatComponent

\section{Additional evaluations}
\label{sec:sub_add_eval}

\subsection{Multi-view to materials - evaluations on material components}

In \cref{tab:mat_comp} and Figure~\ref{fig:mat_component} we evaluate the quality of the individual material components. As the geometry and hence UV-space is different for LSRM, we compute metrics on g-buffer renderings of the material maps on 17 views of each of the 32 scenes. To compensate for intensity shifts in the reconstructed base color, we additionally report a scale-invariant siPSNR score for that component: given a prediction $x$ and a corresponding reference $y$, we normalize the prediction according to $\hat{x} = x \cdot  \bar{y} / \bar{x}$, 
where $\bar{x}$ represents a tuple with the averages over each image channel.

\subsection{Casual captures}
\figCasual
In \cref{fig:casual} we show material reconstruction from casually captured photographs from the DTC~\citep{dong2025dtc} dataset. Similar to the high quality DTC results from the main paper, we use the high quality geometry from the DTC dataset, but the footage used for material reconstruction comes from a lower quality handheld camera video stream with less accurate poses. We randomly sample 17 frames (using temporal stratified random sampling) from the video stream.

\subsection{Comparison against TEXGen (on diffuse textures)}

TEXGen~\citep{yu2024texgen} introduced generative diffuse material creation in texture space, guided by an additional
3D structure to capture object-space locality. Our approach generates full materials using texture space diffusion
from finetuned 2D diffusion priors. For a fair comparison, we compute metrics for our method and TEXGen in the single-image to material setting
for \textbf{diffuse-only} renderings in \cref{fig:texgen_main} and show a few example generated albedo maps in \cref{fig:texgen}.
Note that our method solves a more complex task: predicting full PBR materials from a shaded view, while TEXGen predicts only albedo maps from a demodulated view. Our method produces more coherent materials and we score higher in all metrics computed on the evaluation set of 2048 relit views.

\figTEXGen

\subsection{DTC material reconstruction results}
In \cref{fig:dtc_scenes} we show further real world results from the DTC dataset. While our method scores slightly lower on view reconstruction than differentiable rendering (DiffPT) we note that the texture details are  more consistent and sharper. 

\figDTC

\subsection{Upscaler} 
Our upscaler network mainly serves as a tool that we use along with noise rolling~\citep{bartal2023multidiff,vecchio2023controlmat} for higher resolution results, and is designed similarly to previous work. Current generative pipelines typically use ESRGAN~\citep{wang2018esrgan} to either directly upscale material parameter textures~\citep{hunyuan3d2025}, or to upscale rendered images and rely on differentiable rendering tricks to propagate the results to the material parameters, one notable example is PBR-SR~\citep{chen2025pbrsr}. We report quantitative results in \cref{tab:upscaler}, as expected our upscaler performs on par with previous work, with differences mainly being due to training dataset and our method being intended for texture space use. Notably PBR-SR~\citep{chen2025pbrsr} performs better on roughness and metallicity parameters, which is due to their expensive multi-view differentiable rendering optimization step. 

\tabUpscaler

\subsection{Impact of 3D-aware RoPE}
\label{sec:sup_rope}

\figAttention

\figRoPEAttnSup
\tabRopeAblation

We ablate our 3D-aware RoPE, combining a 3D world space position and frame id: $\mathbf{p}_{xyz} + f_\mathrm{id}$, against the standard Wan~2.1 RoPE (combining pixel $xy$-coordinates, $\mathbf{p}_{uv}$ and $f_{\mathrm{id}}$) on the text to material application, where the impact is largest. 
We ablate against both the standard Wan~2.1 RoPE, and Wan~2.1 RoPE+$G_{\mathrm{buf}}$ where the world space position and normals are provided as VAE encoded guide images. 
In \cref{fig:attention} we show attention maps for a particular query point at selected layers at diffusion denoising step 15 (out of 50). As expected, our 3D-aware~RoPE attends more strongly to local regions in world space, even if they are distant in texture space. 
In \cref{fig:rope_attn_supplemental}, we show the attention maps for all layers, and note that this behavior is apparent through all layers of the network. The Wan~2.1 RoPE+$G_{\mathrm{buf}}$ variant shows similar behavior but only in the middle layers of the network. \cref{tab:rope_ablation_sup} ablates the RoPE variants on text to material generation for our full synthetic dataset. Our 3D-aware RoPE shows a small but consistent improvement in quantitative metrics. 
This ablation uses the same model architecture and parameter count as the main experiments, but generates material textures at $512 \times 512$ rather than $2048 \times 2048$ resolution, which is why the results do not completely match the main evaluation in \cref{tab:matgen}. \cref{fig:figRoPE} shows an additional example scene from our synthetic dataset, where we run the same text to material generation with four different seeds. The Wan~2.1 RoPE results show higher variance, with the model struggling to separate geometric components in texture space.

\figRoPE

\paragraph{Text to Material.} 
\figTextToMaterial
In \cref{fig:text_to_material} we show examples of materials generated from text conditioning only. Even though our method operates in disjoint texture space, the 3D RoPE encoding provides enough information to generate semantically meaningful materials. Texture space discontinuities also often coincide with material borders (e.g. the ``Marshall'' text) which can sometimes benefit our model over the image space alternatives.

\clearpage

\end{document}